\documentclass[11pt]{article}

\usepackage{acl}

\usepackage[utf8]{inputenc}
\usepackage{csquotes}
\usepackage{natbib}

\usepackage{times}
\usepackage{latexsym}

\usepackage{booktabs}
\usepackage{rotating}
\usepackage{pdflscape}
\usepackage[T1]{fontenc}
\usepackage{pdfpages}
\usepackage{float}

\usepackage{hyperref}
\usepackage{microtype}
\usepackage{glossaries} 

\usepackage{tabularx}
\usepackage{multirow}
\usepackage{amssymb} 
\usepackage{pifont} 

\usepackage{xcolor}
\definecolor{darkergreen}{RGB}{0,120,0} 
\definecolor{darkerred}{RGB}{160,0,0} 

\usepackage{ragged2e}

\usepackage{setspace}

\usepackage{graphicx} 
\usepackage{epigraph}

\usepackage{array} 
\newcolumntype{L}[1]{>{\raggedright\let\newline\\\arraybackslash\hspace{0pt}}p{#1}}
\newcolumntype{H}{>{\setbox0=\hbox\bgroup}c<{\egroup}@{}}

\usepackage{tikz}

\title{S\textsuperscript{3} - Semantic Signal Separation
}

\author{
  Márton Kardos \\
  Aarhus University\\ 
  \texttt{\small martonkardos@cas.au.dk} \\\And
  Jan Kostkan \\
  Aarhus University \\
  \texttt{\small jan.kostkan@cas.au.dk} \\\And
  Kenneth Enevoldsen \\
  Aarhus University \\
  \texttt{\small kenneth.enevoldsen@cas.au.dk} \\\AND
   Arnault-Quentin Vermillet \\
  Aarhus University \\
  \texttt{\small arnault@cc.au.dk}  \\\And
  Kristoffer Nielbo \\
  Aarhus University \\
  \texttt{\small kln@cas.au.dk} \\\And
  Roberta Rocca \\
  Aarhus University \\
  \texttt{\small roberta.rocca@cas.au.dk}
}

\begin{document}
\maketitle
\begin{abstract}
Topic models are useful tools for discovering latent semantic structures in large textual corpora.
Recent efforts have been oriented at incorporating contextual representations in topic modeling and have been shown to outperform classical topic models.
These approaches are typically slow, volatile, and require heavy preprocessing for optimal results.
We present \mbox{\textit{Semantic Signal Separation}} (S\textsuperscript{3}), a theory-driven topic modeling approach in neural embedding spaces.
S\textsuperscript{3} conceptualizes topics as independent axes of semantic space and uncovers these by decomposing contextualized document embeddings using \textit{Independent Component Analysis}.
Our approach provides diverse and highly coherent topics, requires no preprocessing, and is demonstrated to be the fastest contextual topic model, being, on average, 4.5x faster than the runner-up BERTopic\footnote{The median runtime ratio of BERTopic and $S^3$ on runs paired by dataset, encoder model and number of topics is 4.5.}.
We offer an implementation of S\textsuperscript{3}, and all contextual baselines, in the \texttt{Turftopic}\footnote{\url{https://github.com/x-tabdeveloping/turftopic}} Python package.
\end{abstract}

\section{Introduction}

`Topic models' are an umbrella term for statistical approaches that enable unsupervised topic discovery in large text corpora \citep{blei_prob_topic_models}.
They are commonly applied in exploratory data analysis of textual data because they allow practitioners to unearth and condense information about the semantic content of a corpus without the need for close reading and manual labor. 
Traditionally, topics are presented to the user as a set of important terms (keywords) that provide insights into possible interpretations of the topic.

\begin{figure}[htbp]
    \centering
    \includegraphics[width=.95\linewidth]{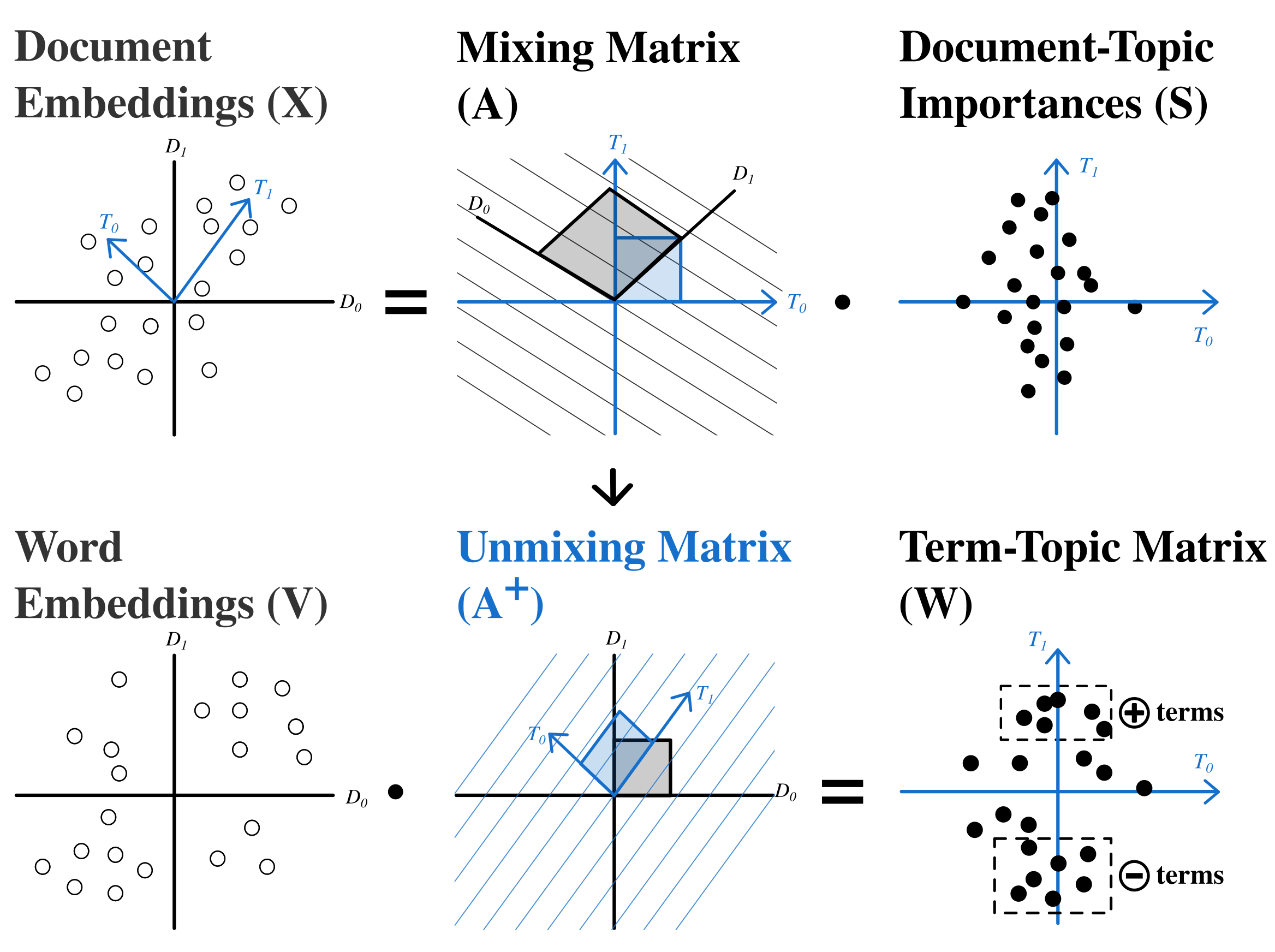}
    \caption{
    S\textsuperscript{3} discovers topics by finding independent semantic axes that explain most variance in an embedded corpus and interprets these semantic axes as topics.
    Descriptive words for a topic are found by projecting word embeddings onto these axes.
    }
    \label{fig:s3_overview}
\end{figure}

\begin{figure*}[t]
    \includegraphics[width=.98\linewidth]{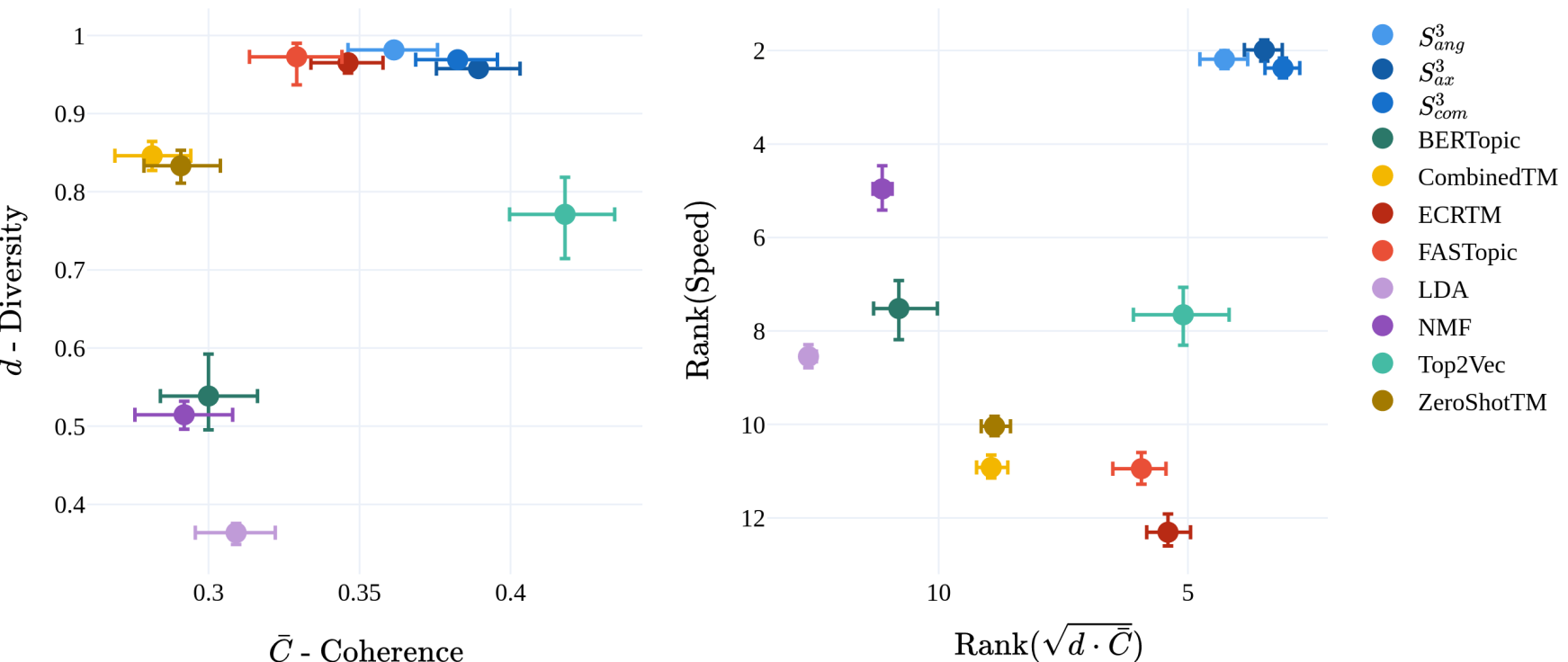}
    \caption{Coherence and diversity of topic descriptions from all topic models (left) 
    and topic models' ranks on speed and performance across all runs (right). \\
    \textit{Error bar represents 95\% bootstrap confidence interval. Results are demonstrated on raw corpora, only using contextual embedding models.}
    }
    \label{fig:ranks}
\end{figure*}

Classical approaches to topic modeling, such as Latent Semantic Indexing (\citealp{lsi_information_retrieval}, \citealp{lsa}) and Latent Dirichlet Allocation (\citealp{blei_lda}, \citealp{blei_prob_topic_models}), have relied on frequency-based bag-of-words (BoW) representations of documents.
While these models have been successfully utilized in decades of NLP research \citep{lda_survey}, they all share several practical and theoretical limitations.
For example, BoW models are sensitive to words with atypical statistical properties (such as function words), which can contaminate keyword-based topic descriptions unless heavy preprocessing pipelines are applied.
Such pipelines introduce many degrees of freedom for the researcher.
Furthermore, the sparsity and high dimensionality of BoW representations often result in lower computational efficiency and poorer model fit.

With the advent of dense neural language representations (\citealp{attention_is_all_you_need}, \citealp{le2014distributed}, \citealp{glove}, \citealp{mikolov2013efficient}), new opportunities have opened for topic modeling research.
Sentence embeddings \citep{sentence_transformers} hold great promise for topic modeling, as 
they provide contextual, grammar-sensitive representations of language,
and are more robust against spelling errors and out-of-vocabulary terms, reducing the need for preprocessing that discards valuable linguistic information. Additionally, they
produce dense representations in a continuous space, allowing for assumptions of Gaussianity.
They also allow for transfer learning \citep{ruder-etal-2019-transfer} in topic modeling, utilizing information learned in larger external corpora for topic extraction.

Several approaches have thus been proposed using dense neural representations for contextual topic modeling,
and these have been shown to outperform their non-contextual counterparts (\citealp{ctm}, \citealp{zeroshot_tm}, \citealp{bertopic_paper}, \citealp{top2vec}, \citealp{fastopic}).
Many of these approaches, however, still require preprocessing for optimal performance.
This is a significant limitation of the field, as preprocessing pipelines are not standardized and can remove valuable information which is especially impactful for shorter texts \cite{wu-etal-2020-short}.

\subsection{Contributions}
We present Semantic Signal Separation (S\textsuperscript{3}),
a novel contextualized topic modeling technique that conceptualizes topic modeling as the discovery of latent semantic axes in a corpus.
These axes are discovered by decomposing the document embedding matrix using the FastICA \citep{ica_fast} algorithm.

The proposed approach is
a) Conceptually simple and theory-driven
b) Performs on par with existing approaches in word-embedding coherence and produces near-perfect diversity
c) Is computationally more efficient than existing approaches, and
d) can effectively utilize contextual information

In addition to introducing a new method, we provide a simple unified scikit-learn based interface for both S\textsuperscript{3} and other contextualized topic modeling approaches in the \texttt{Turftopic}  Python package \footnote{\url{https://github.com/x-tabdeveloping/turftopic}}.

\section{Related Work}

\subsection{Semantic Axes}

Independent Component Analysis \citep{ica_original} has previously been applied to embedding spaces to discover semantic axes (\citealt{musil-marecek-2024-exploring}, \citealt{yamagiwa-etal-2023-discovering}).
These investigations are, however, mostly oriented at finding universal dimensions of semantics in word and image embeddings.
They demonstrated that axes discovered by ICA are interpretable and usually coincide across different embedding spaces and modalities.
In contrast, our study is oriented at utilizing semantic axes to discover highly interpretable topics in a specific corpus of interest, not at uncovering universal dimensions of semantics.
Additionally, no topic descriptions are computed or evaluated in these studies.

\subsection{Embedding-based Topic Models}
Multiple approaches to topic modeling using neural language representations have been proposed over the past few years.

\textbf{Neural Topic Models} \citep{neural_survey} rely on deep neural networks for parameter estimation.
Contextualized Topic Models or CTMs \citep{ctm} are generative models of BoW representations,
but use a variational autoencoding paradigm for inference \citep{prodlda}. 
Contextual embeddings are used as inputs to the encoder network (\textit{ZeroShotTM}) at times concatenated with BoW vectors (\textit{CombinedTM}).
CTMs typically require heavy preprocessing,
and computational efficiency and quality of model fit decrease drastically with larger vocabularies \citep{ctm_docs}.

ECRTM \citep{ecrtm} is a neural model that relies on embedding clustering regularization to produce sufficiently distinct topics and prevent topic descriptions from converging to each other.
This approach, however, does not utilize contextual representations and is notably slower than most other topic models \citep{fastopic}.

FASTopic \citep{fastopic} introduces a dual-semantic-relation paradigm where relations between documents, topics, and words are conceptualized as optimal transport plans.
As they demonstrate, their approach is more efficient and produces higher-quality topics than previous neural approaches.

\textbf{Clustering Topic Models} discover topics in corpora by clustering document representations in embedding space.
Word importance weights for a given topic are estimated post hoc.

Top2Vec \citep{top2vec} estimates these by computing cosine similarity between word encoding and cluster centroids.
This assumes clusters to be spherical and convex, and topic descriptions might be misrepresentative depending on cluster shape.

BERTopic \citep{bertopic_paper} estimates term importances for clusters using a class-based tf-idf weighting scheme.
Both approaches utilize UMAP \citep{umap_paper} for dimensionality reduction and HDBSCAN \citep{hdbscan} for clustering.

Both BERTopic and Top2Vec come pre-packaged with a \textit{topic reduction} method.
This is necessary, as HDBSCAN learns the number of clusters from the data and the number of clusters can grow vast, which may prove impractical.

\textbf{The Challenges} of currently available contextual topic models are, however, still numerous.
Many of them are sensitive to hyperparameter choices, produce topics of dubious interpretability, and rely on preprocessing pipelines
the structure of which is not standardized \citep{neural_survey}.
Additionally, it is unclear whether these models are effective at using contextual and syntactic information, as they are typically evaluated on preprocessed corpora.

\section{Semantic Signal Separation}
In this paper, we introduce \textit{Semantic Signal Separation} (or \textbf{S\textsuperscript{3}}), a novel approach to topic modeling in continuous embedding spaces that aims to overcome the above-mentioned challenges of existing contextual topic modeling methods.

Instead of interpreting topics as clusters or word probabilities, we conceptualize topics as semantic axes that explain variation specific to a corpus.
This is achieved by decomposing semantic representations into latent components ($A$), which are assumed to be the topics, and components' strengths in each document ($S$), which are document-topic importances.
For the topics to be conceptually distinct, we utilize Independent Component Analysis \citep{ica_original} to uncover them.
Term importances are estimated from the topic components' strength in word embeddings ($V$).

Our approach can, in some aspects, be considered the contextual successor of Latent Semantic Analysis (\citealp{lsa}, \citealp{lsi_information_retrieval}),
which discovers factors in word-occurrences.

\subsection{Model}

\textbf{Document representations} are obtained by encoding documents using a sentence transformer model.

\begin{enumerate}
    \item Let $X$ be the matrix of the document encodings.
\end{enumerate}

\noindent
\textbf{Decomposition} of document representations into independent semantic axes is performed with Independent Component Analysis \citep{ica_original}.
In this study, we used the FastICA \citep{ica_fast} algorithm to identify latent semantic components.
As a preprocessing step, whitening is applied to the embedding matrix, as FastICA is a noiseless model.
Since ICA discovers the same number of components as the dimensionality of the embeddings by default,
we reduce the dimensionality of embeddings during whitening by taking the first $N$ principal components, where $N$ is the number of topics.

\begin{enumerate}
    \setcounter{enumi}{1}
    \item Decompose $X$ using FastICA: $X = AS$ where $A$ is the mixing matrix, and $S$ the source matrix, containing document-topic-importances.
\end{enumerate}

\noindent
\textbf{Term importances} for topics, which are needed for the selection of descriptive terms, are calculated by projecting words onto the discovered semantic axes.

\begin{figure}[htbp]
    \centering
    \includegraphics[width=.85\linewidth]{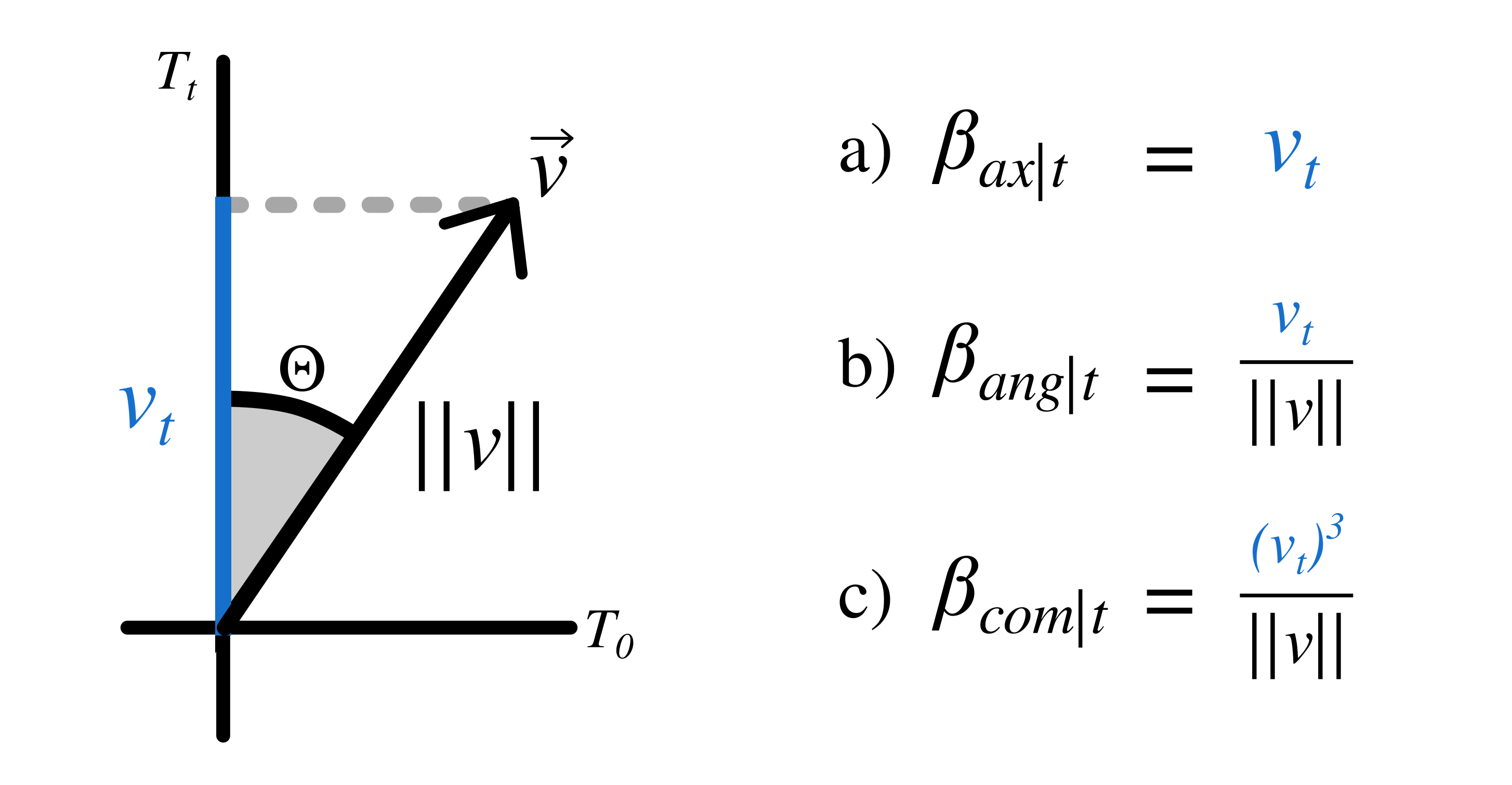}
    \caption{
    Geometric intuition for different types of word importance scores for S\textsuperscript{3}.
    }
    \label{fig:term_importance}
\end{figure}

\begin{enumerate}
    \setcounter{enumi}{2}
    \item Encode the vocabulary of the corpus with the same encoder model.
    Let the matrix of word encodings be $V$
    \item Let the unmixing matrix be the pseudo-inverse of the mixing matrix: $C = A^+$.
    \item Project words onto the discovered semantic axes by multiplying word embeddings with the unmixing matrix: $W = VC^T$.
    \item Calculate word importance scores for each topic.
\end{enumerate}

We examine three methods for computing word importances:
\begin{enumerate}
    \item \textit{Axial} word importances are defined as the words' positions on the semantic axes. The importance of word $j$ for topic $t$ is: $\beta_{tj} = W_{jt}$.
    \item \textit{Angular} topics can be calculated by taking the cosine of the angle between projected word vectors and semantic axes:
    $\beta_{tj} = cos(\Theta) = \frac{W_{jt}}{||W_j||}$
    \item \textit{Combined} word importance is a combination of the two approaches \footnote{We take an odd power of the word position to maintain its sign.}.
    $\beta_{tj} = \frac{(W_{jt})^3}{||W_j||}$
\end{enumerate}

Axial word importance gives rise to topic descriptions that contain the \textit{most salient} words for a given topic, while angular importance weights the \textit{most specific} words highest. Combined importances intend to balance these two aspects.

Note that all formulations allow for terms with \textit{negative} importance for a given topic.
While this is also the case for LSA, prior literature does not explore this concept.
Model interpretation can be augmented by inspecting terms that score lowest on a given topic, providing a \textit{negative definition}.

To ensure comparability with methods that do not allow for negative definitions,
our model comparisons ignore negative terms, but a demonstrative example is presented in Section \ref{sec:arxiv_ml}.

\vspace{0.3cm}
\noindent
\textbf{Inference} of topic proportions in novel documents can be achieved by multiplying the documents' embeddings with the unmixing matrix.
\begin{enumerate}
    \item Let the encodings of previously unseen documents be $\hat{X}$ 
    \item Calculate document-topic matrix: $\hat{S} =  \hat{X} C^T$
\end{enumerate}

\section{Experimental Setup}

To compare S\textsuperscript{3}'s performance to previous context-sensitive topic modeling approaches, we benchmark it on a number of quantitative metrics widely used in topic modeling literature.
Our experimental results can be reproduced with the \texttt{topic-benchmark} Python package's CLI\footnote{\url{https://github.com/x-tabdeveloping/topic-benchmark}}.
The repository also contains results and all topic descriptions in the \texttt{results/} directory.

\subsection{Datasets}

Due to its relevance in topic modeling and NLP research, we benchmark the model on the 20Newsgroups dataset along with
a BBC News dataset\footnote{\url{https://huggingface.co/datasets/SetFit/bbc-news}}, a set of 2048 randomly sampled Machine Learning abstracts from ArXiv,
medical terms' articles from Wikipedia \footnote{\url{https://huggingface.co/datasets/gamino/wiki_medical_terms}}
and StackExchange entries \footnote{\url{https://huggingface.co/datasets/mteb/stackexchange-clustering-p2p}}.
Code is supplied for fetching and subsampling the datasets in a reproducible manner in the \texttt{topic-benchmark} repository.
Consult Table \ref{tab:datasets_encoders} for dataset and vocabulary size.
Topic model comparisons typically use preprocessed datasets from the OCTIS Python package \citep{octis_paper}.
To estimate the effect of preprocessing on model performance, we run model evaluations on the preprocessed version of 20Newsgroups as well, and compare results with the non-preprocessed dataset.

\begin{table}[h]
\resizebox{0.5\textwidth}{!}{
\begin{tabular}{lll}
\toprule
\textbf{Dataset} & \textbf{\# Documents} & \textbf{Vocabulary Size} \\
\midrule
ArXiv ML Papers & 2048 & 2849\\
BBC News & 1225 & 3851\\
20 Newsgroups Preprocessed & 16310 & 1612\\
20 Newsgroups Raw & 18846 & 21668\\
StackExchange & 75000 & 17884\\
Wiki Medical & 6861 & 22145\\
\toprule
\textbf{Embedding Model} & \textbf{\# Parameters} & \textbf{Embedding Size} \\
\midrule
Averaged GloVe & 120 M & 300 \\
all-MiniLM-L6-v2 & 22.7 M & 384 \\
all-mpnet-base-v2 & 109 M  & 768 \\
E5-large-v2 & 335 M & 1024 \\
\bottomrule
\end{tabular}
}

\caption{Overview of Datasets and Embedding Models}
\label{tab:datasets_encoders}
\end{table}

\subsection{Embedding Models}
To evaluate the effect of embedding models on performance, we run all analyses with an array of embedding models of varying size and quality.
We utilized two SBERT models \citep{sentence_transformers}, a static GloVe model averaging word vectors \citep{glove}, as well as an E5 model \citep{e5} (see Table \ref{tab:datasets_encoders}). 

\subsection{Baseline Models}
Topic models included in the experiment were BERTopic, Top2Vec, ZeroShotTM, CombinedTM, FASTopic, and ECRTM along with two classical baselines: NMF and LDA.

For a detailed discussion of hyperparameters, consult Appendix \ref{sec:hyperparameters}.

All models were run using 10, 20, 30, 40, and 50 topics \footnote{Similar to \citet{bertopic_paper}, the number of topics was reduced to the desired amount in Top2Vec and BERTopic},
and the topic descriptions from each model (top 10 terms in each topic) were extracted for quantitative and qualitative evaluation. 
Topic models with a given number of topics were only run once, due to the large computational demands of estimating topics models for the full battery on model types, encoders, and datasets.

\begin{figure*}[t!]
    \makebox[\textwidth]{
    \includegraphics[width=1.0\linewidth]{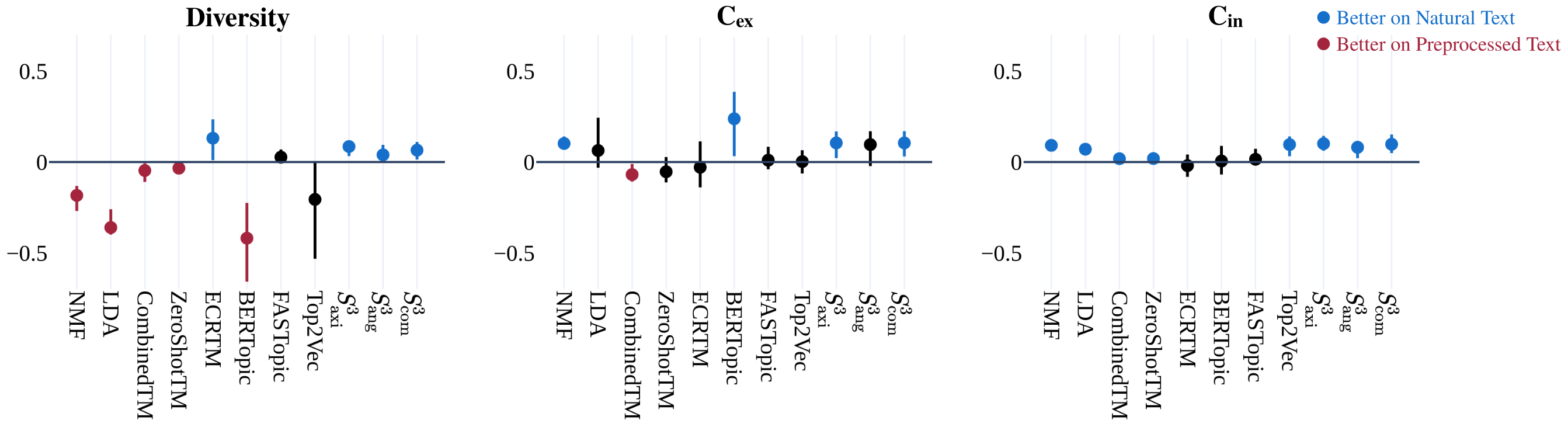}}
    \caption{Performance difference between topic models on natural text and preprocessed data (20 Newsgroups) \\
    S\textsuperscript{3} is the only model that consistently performs better on natural text. \\
    \textit{Error bars represent 95\% highest density interval.} \\
    }
    \label{fig:effect_of_preprocessing}
\end{figure*}

\subsection{Metrics of Topic Quality}
We evaluate the quality of 10-word topic descriptions in terms of topic diversity and coherence, using the following metrics.

\textbf{Topic diversity}($d$)  measures how different topics are from each other based on the number of words they share.
Topic diversity is essential for interpretability: when many topics have the same descriptive terms, it becomes hard to delineate their meaning.\citep{etm_paper}

\textbf{Topic coherence}($C$) measures how semantically coherent topics are.
In our investigations, we used word embedding coherence \citep{ctm},
which equates to the average pairwise similarity of all words in a topic description, based on a Word2Vec model.
Typically, models, which have been pre-trained on large corpora of text are used, which capture words' semantic similarity in general.
\footnote{We used the \texttt{word2vec-google-news-300} word embedding model \citep{gensim}.}
Since these semantic relations are not specific to the corpus studied, we will refer to this approach as \textit{external} coherence ($C_{\text{ex}}$).
It is, however also beneficial to gain information about \textit{internal} coherence ($C_{\text{in}}$), that is, how well a topic model captures semantic relations between words in a specific target corpus.
As such, we also computed coherence using Word2Vec models trained on the corpus, from which the topics were extracted.
To gain an aggregate measure of topic coherence, we also utilized the geometric mean of these approaches: 
$\bar{C} = \sqrt{C_{\text{ex}} \cdot C_{\text{in}}}$

In addition, when an aggregate measure of topic interpretability is needed, we took the geometric mean of coherence and diversity
$\sqrt{\bar{C} \cdot d}$.
\footnote{We used geometric, instead of arithmetic mean, as it is better at capturing \textit{both $a$ \textbf{and} $b$} types of relations. For instance, topics with 1.0 coherence and 0.0 diversity, should ideally not get a score of 0.5 as this would suggest reasonable performance.}


\subsection{Metrics of Robustness}

With classical topic models, it is fairly common that semantically irrelevant ``junk terms'', contaminate topic descriptions. 
Ideally, a topic model should yield topic descriptions where only terms aiding the interpretation of a given topic are present.
This core property of topic models is not captured by standard evaluation metrics. 

For each topic model fitted on the raw corpus, we computed the relative frequency of \textbf{stop words} in topic descriptions.
As a proxy for the prevalence of ``junk terms'', we also computed how frequently \textbf{nonalphabetic characters} are part of terms included in topic descriptions. Note that this is not a perfect proxy for the meaningfulness of terms: non-alphabetical terms might enhance topic descriptions under certain circumstances (e.g., "1917" is a meaningful term in a topic about the October Revolution).

\section{Results}

Evaluations demonstrate that S\textsuperscript{3} is substantially more balanced than baselines,
generally outperforming them on aggregate performance,
and, on average, being 27.5x faster\footnote{Median runtime ratio on runs paired by dataset, embedding model and number of topics against all baselines} than the baselines.
Our models rank highest both on aggregate performance, but also quite consistently in runtime (see Figure \ref{fig:ranks}). 
S\textsuperscript{3} performed consistently well across corpora and embedding models, and was only occasionally rivaled by Top2Vec, ECRTM and FASTopic.
On average, ECRTM and FASTopic resulted in more diverse, but less coherent topics, while Top2Vec resulted in highly coherent, but less diverse topics.
In contrast, S\textsuperscript{3} strikes an optimal balance of diversity and coherence.
To see disaggregated results and runtimes, consult Figures \ref{fig:disagg_coherence}, \ref{fig:disagg_diversity} \ref{fig:disagg_wec_ex}, \ref{fig:disagg_wec_in} and \ref{fig:disagg_runtime}.

In order to determine whether differences in model performance were significant, a linear regression analysis was performed.
We predicted the aggregate interpretability score ($\sqrt{\bar{C} \cdot d}$), with topic model type as a fixed effect (with $S^3_{com}$ as the intercept), and the number of topics, encoder model and dataset as random intercepts.
It was found that model type significantly predicts interpretability ($F = 167.4$; $p<0.001$; $R^2 = 0.673$), and all models' (except for the other two variants of $S^3$) coefficients were negative and were significant (all $p < 0.05$), meaning that $S^3$ significantly outperformed all other models on interpretability.
We report the full table of coefficients in Table \ref{tab:regression}.

\subsection{Effects of Preprocessing}

To test how sensitive models are to preprocessing, results on the raw, and preprocessed 20 Newsgroups datasets were compared.
While some baselines also improved in coherence metrics when having access to the raw corpus, their performance was, at large, identical or worse without preprocessing.
S\textsuperscript{3} variants gained by far the most from removing preprocessing, indicating that they are effective at utilizing the additional information (see \ref{fig:effect_of_preprocessing})
This is also indicated by the fact, that, while S\textsuperscript{3} was outperformed by some baselines when heavy preprocessing was applied,
its performance on the raw corpus was higher than all other models,
even the ones trained on the preprocessed data.

\subsection{Stop Words}

\begin{figure}[h]
    \includegraphics[width=.85\linewidth]{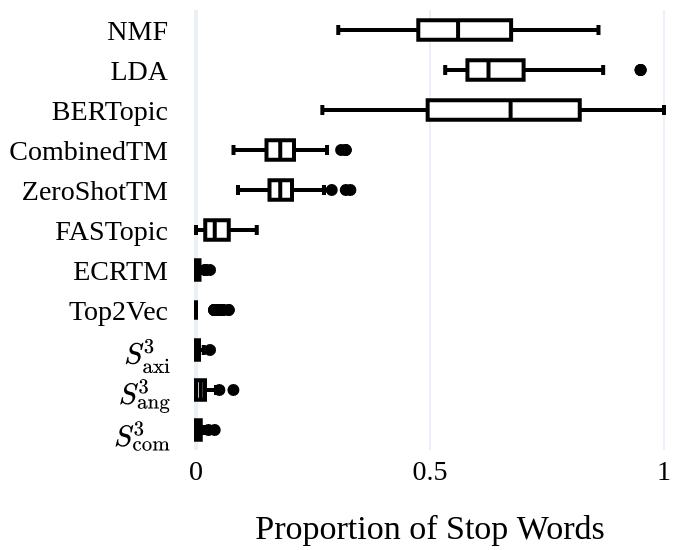}
    \caption{Relative frequency of stop words in topic descriptions}
    \label{fig:stop_words}
\end{figure}

When evaluated on raw corpora, many topic models included quite a few stop words in the top 10 words.
This tendency was especially prevalent with BoW models and BERTopic, at times making up 100\% of topic descriptions.
CTMs and FASTopic performed considerably better, while ECRTM, Top2Vec and all variants of S\textsubscript{3} resulted in topic descriptions rarely containing stop words.

We did not observe any patterns with nonalphabetical characters, most models performed quite similarly in this aspect (see Figure \ref{fig:nonalphabetical}).

\subsection{Effects of Embedding Models}

\begin{figure}[ht]
    \includegraphics[width=\linewidth]{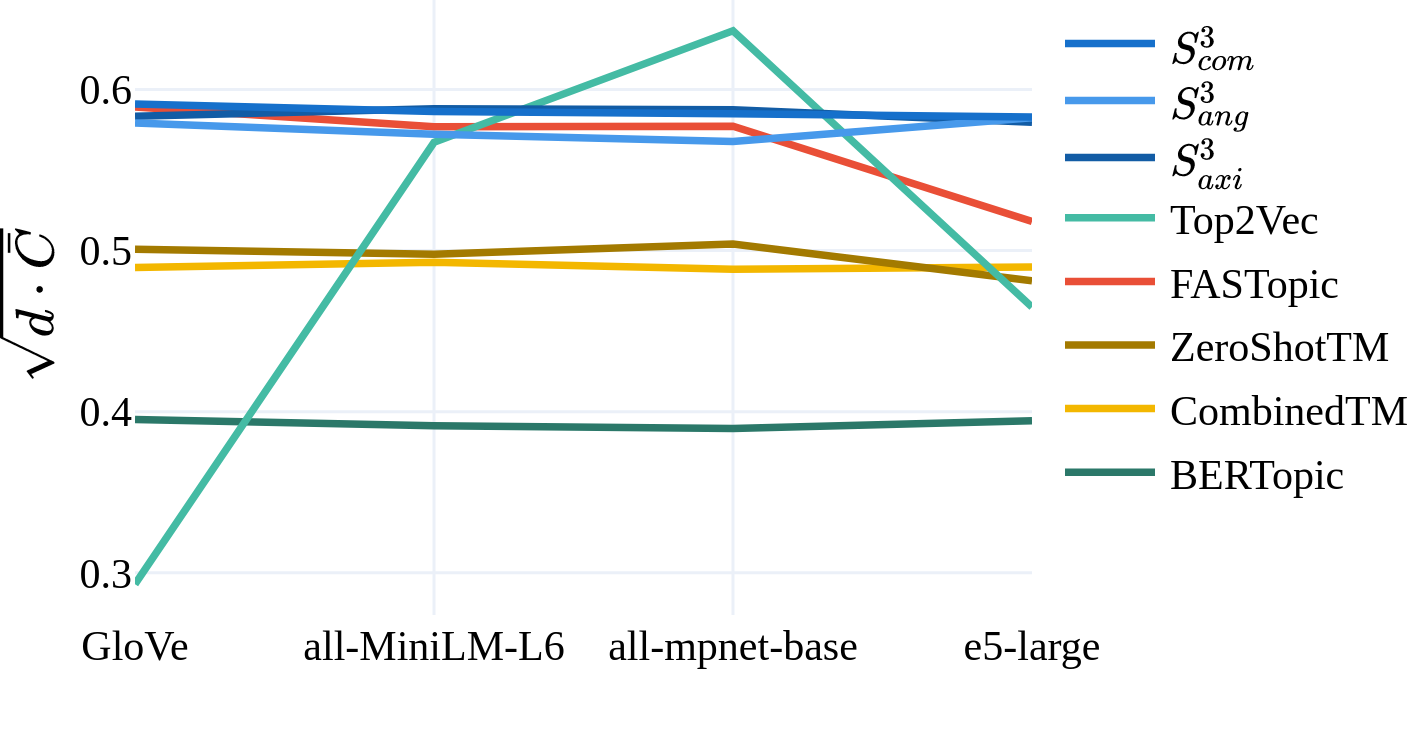}
    \caption{Effect of embedding model on contextual topic models' aggregate performance.}
    \label{fig:effect_of_embedding_model}
\end{figure}

While most resultes reported were produced with representations from sentence transformers,
we also examined performance on averaged static word-embeddings.
Being able to utilize these representations might prove useful in low-resource environments.
Generally, S\textsuperscript{3} performed relatively stably across embedding models.
By far, the most affected model was Top2Vec, which performed drastically worse with GloVe and E5 embeddings.
FASTopic, by contrast, performed best with GloVe embeddings, and its performance dropped with each increase in embedding model size (see Figure \ref{fig:effect_of_embedding_model}).
This is likely due to the fact that the model suffers from the curse of dimensionality, and the quality of fit decreases with embedding size.

\subsection{Effects of Term Importance Estimation}
Evaluations were run with all variants of S\textsuperscript{3}.
The performance of all three term importance estimation methods was relatively similar, with differences generally representing a trade-off between coherence and diversity.
Angular term importance resulted in topics being more diverse, while axial topics were the most coherent.
The combined method generally splits this difference.
We observed that the combined method regularly resulted in topics that differed only in a couple of words from their axial counterpart.
This suggests that highly relevant words are also usually specific enough to a given axis.
As such, we recommend that one uses the \textit{combined} method by default, as it can prevent a word penetrating multiple topic descriptions at the same time, when it scores high on multiple axes.

\section{Qualitative Evaluation}

\subsection{Model Comparison}
We inspect topics qualitatively to identify whether patterns observed in quantitative metrics reflect intuitive characteristics of model output. For feasibility, we will focus on the models (among those evaluated above) which extract 20 topics from the 20 Newsgroups dataset.
To see all topic descriptions, consult Appendix \ref{sec:topic_desc}.

In line with previous analyses, LDA, NMF and BERTopic resulted in the least interpretable topics.
Topic descriptions often consisted of function words and acronyms.
Even topics which contained information-bearing words were in most cases hard to interpret: 
\begin{itemize}
    \item \textit{that, to, you, of, from, the, and, in, was, on} (LDA)
    \item \textit{the, of, to, in, space, it, edu, is, that, and} (BERTopic/e5-large-v2)
\end{itemize}

CTMs and ECRTM performed notably better, with a lot of topics being readily interpretable. Yet, many of the topics in these models were still hard to interpret:

\begin{itemize}
    \item \textit{145, ax, 0d, \_o, a86, mk, m3, mp, 0g, mm} (CombinedTM/all-MiniLM-L6-v2)
    \item \textit{verbeek, billington, cassels, c5ff, nyr, det, bos, guerin, nieuwendyk, ashton} (ECRTM)
    \item \textit{get, good, my, car, doctor, diet, patients, ve, too, like} (ZeroShotTM/e5-large-v2)
\end{itemize}

FASTopic's topics were even more informative, and contained less noise, but occasionally conflated two conceptually distinct topics into one.

\begin{itemize}
    \item \textit{miles, dealer, auto, engine, ford, oil, cars, honda, toyota, mustang} (FASTopic/all-mpnet-base-v2)
    \item \textit{moon, launch, henry, bike, medical, car, dod, orbit, shuttle, mission} (FASTopic/all-mpnet-base-v2)
\end{itemize}

The most specific, most informative and cleanest topics were produced by Top2Vec and S\textsuperscript{3}, with most topics being informative and intuitively understandable, as showcased by these examples:

\begin{itemize}
\item \textit{malpractice, diagnosis, doses, homeopathy, medical, diagnoses, poisoning, toxins, gastroenterology, biomedical} (Top2Vec/all-mpnet-base-v2)
\item \textit{epilepsy, medical, toxins, medicines, malpractice, resurection, diseases, homeopathy, poisoning, remedies} ($S^3_{axi}$/all-mpnet-base-v2)
\item \textit{zionists, israelis, israeli, intifada, zionist, israel, palestinians, likud, palestinian, isreal} ($S^3_{axi}$/e5-large-v2)
\end{itemize}

In line with our quantitative evaluations, it can be observed that some models, especially Top2Vec, are negatively affected by E5 embeddings.
In contrast, S\textsuperscript{3} produced the highest quality topics with E5 embeddings, indicating that the model can effectively utilize representations of higher quality.
Models were also generally negatively impacted by using non-contextual GloVe embeddings.
S\textsuperscript{3} still performed reasonably well with these. It did,
however, include more noise in topic descriptions than with contextually sensitive embedding models.
FASTopic was virtually unaffected by using non-contextual embeddings.

\subsection{Demonstration: Semantic Axes in ArXiv ML Papers}
\label{sec:arxiv_ml}
As previously mentioned, S\textsuperscript{3} is capable of providing negative descriptions of topics, by extracting the lowest-scoring terms on the given topic.
To demonstrate S\textsuperscript{3}'s unique capabilities to describe semantic variation in a corpus,
we extracted five topics from the same subset of ArXiv ML Papers as were used for quantitative evaluations.
See Table \ref{tab:arxivml} for the top 5 positive and negative terms in each topic.

Note that in this case, we gain notably more information about a topic by inspecting negative terms as well.
Topic 0 and 4 resemble each other quite a bit in positive descriptive terms, and it is only a glance at the negative terms that clarifies where the difference lies between them.

\begin{figure}[htbp]
    \centering
    \includegraphics[width=.9\linewidth]{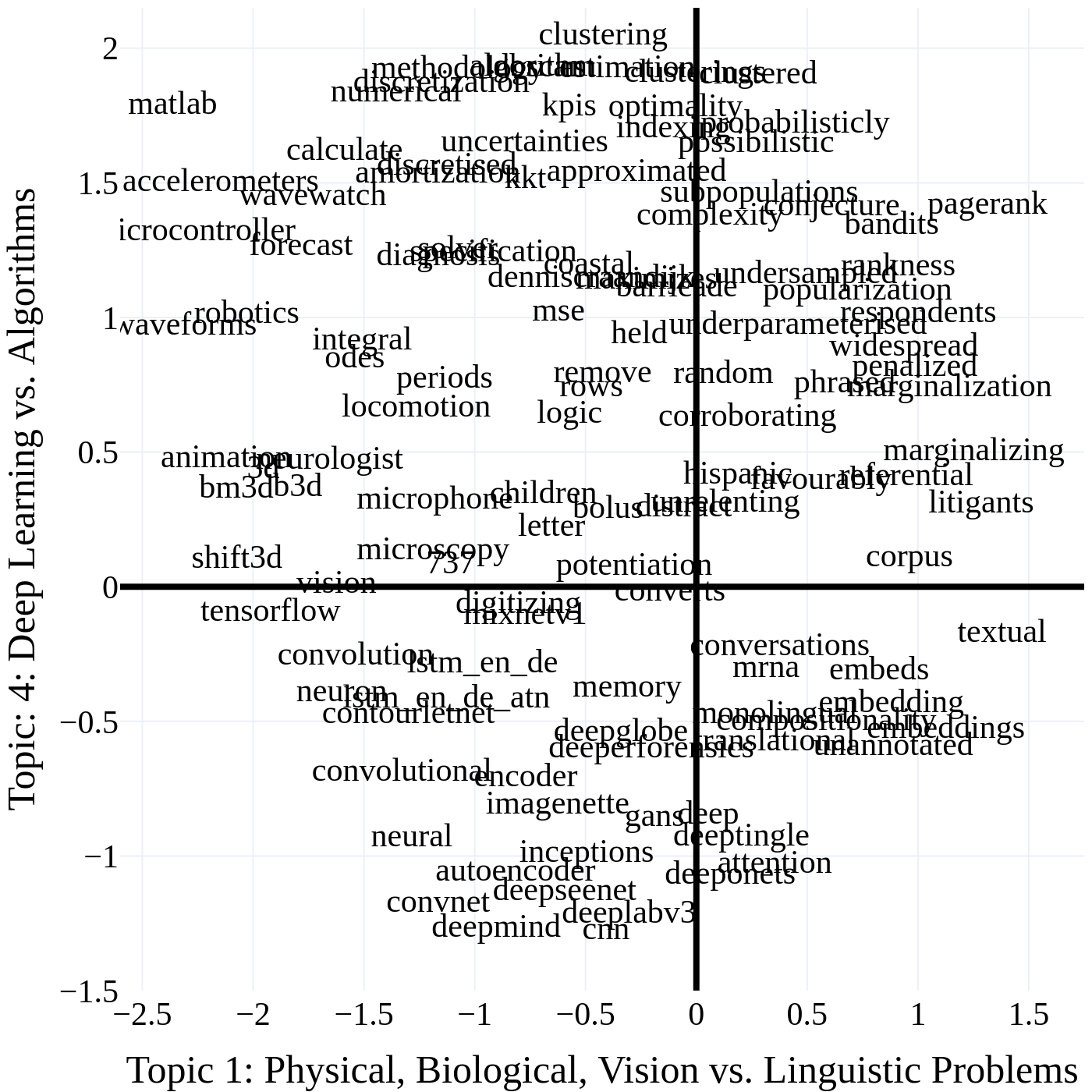}
    \caption{Concepts along Semantic Axes Extracted by S\textsuperscript{3} in ArXiv ML Papers}
    \label{fig:arxiv_word_map}
\end{figure}

S\textsuperscript{3} can also be used to create a \textit{concept compass} of terms along two chosen axes.
This allows us to study concepts in a corpus along the axes extracted by the model, thereby gaining information about the axes' interactions.
We chose Topic 1, which seems to be about \textit{Linguistic vs. Physical/Biological/Vision} problems, and Topic 4, which seems to be about \textit{Algorithmic vs. Deep Learning} solutions. (See Figure \ref{fig:arxiv_word_map})

We can see that the intersection of linguistic problems and classical ML is largely dominated by probability theory, and the term that is quite high on both axes is \textit{pagerank}, which is an algorithmic method for calculating page importance, while deep-learning in linguistic problems is focused on \textit{embeddings} and \textit{attention}.
For low-level problems on the algorithmic side of the spectrum are numerical methods, \textit{matlab} and sensors, while on the deep learning side, are terms related to computer vision and \textit{tensorflow}.

\begin{table}[h]
\footnotesize
\centering
    \begin{tabularx}{\linewidth}{ p{0.05cm} >{\RaggedRight} X >{\RaggedRight} X}
    \toprule
    & \textbf{Positive} & \textbf{Negative} \\
    \midrule
0 & clustering, histograms, clusterings, histogram, classifying & reinforcement, exploration, planning, tactics, reinforce\\
1 & textual, pagerank, litigants, marginalizing, entailment & matlab, waveforms, microcontroller, accelerometers, microcontrollers\\
2 & sparsestmax, denoiseing, denoising, minimizers, minimizes & automation, affective, chatbots, questionnaire, attitudes\\
3 & rebmigraph, subgraph, subgraphs, graphsage, graph & adversarial, adversarially, adversarialization, adversary, security\\
4 & clustering, estimations, algorithm, dbscan, estimation & cnn, deepmind, deeplabv3, convnet, deepseenet\\
    \bottomrule
    \end{tabularx}
\medskip
\caption{Topics in ArXiv ML Papers Extracted by S\textsuperscript{3}}
\label{tab:arxivml}
\end{table}

\section{Conclusion}
We propose Semantic Signal Separation (S\textsuperscript{3}), a novel approach to topic modeling in continuous semantic spaces.
Inspired by classical matrix decomposition methods, such as Latent Semantic Analysis,  S\textsuperscript{3} conceptualizes topics as axes of semantics.
Through experimental quantitative and qualitative evaluation,
we demonstrate that S\textsuperscript{3} discovers highly coherent and diverse topics,
performs well, and in fact, better without preprocessing,
and is, on average, faster than existing contextual topic models, while obtaining better or comparable topics in terms of coherence and diversity.




\section{Limitations}

\subsection{Quantitative Metrics}
While adopted by the majority of topic modeling literature (\citealp{ctm}, \citealp{bertopic_paper}), metrics commonly used to measure topic quality rely on strong assumptions, and they are affected by a number of limitations \citep{rahimi-etal-2024-contextualized}. Additional analyses and considerations presented in this work are meant to partly compensate for this limitation.

\subsection{Model Implementations}
As alluded to, all contextual models we used in the experiment were reimplemented as part of the \texttt{Turftopic} package.
Our implementations of BERTopic and Top2Vec should behave identically to the original, as they rely on the exact same algorithm.
Minor deviances in inference speed (in either direction) are, however, possible.
On the other hand, CTM models posit minor architectural differences to the original implementation, and as such, runtimes and topic descriptions may be slightly different from what we could have obtained with the original implementation.

\subsection{Hyperparameter Tuning}
Some baselines, such as BERTopic and LDA are widely known to be sensitive to choice of hyperparameters.
These approaches could, in theory, perform better with hyperparameter optimization.
We do not tune hyperparameters when comparing models for reasons outlined in Appendix \ref{sec:hyperparameters}.

\subsection{Stochastic Experiments}
Since some of the experiments conducted in the paper are of a stochastic nature, it would have made our analysis more robust to run these using multiple random seeds.
We only ran these with a single seed due to the evaluation-pipeline's runtime.
We consider running topic models with a multitude of embedding models to somewhat compensate for this limitation.

\subsection{Preprocessing Effects}

Most comparable literature does not account for the effects of their preprocessing pipeline.
We have conducted an experiment of this aspect of models on the 20 Newsgroups corpus.
The results presented in this paper could be made more robust by extending this evaluation to multiple corpora.

\subsection{Evaluating Document-Topic Proportions}

While most relevant literature evaluates document-topic proportions as document representations for downstream tasks, such as classification or clustering,
we have decided not to do this,
as we believe, that NLP practitioners would normally use sentence embeddings for these tasks.
While evaluating the interpretability of these representations on human subjects, should be done in the future.

\bibliography{anthology,custom}
\bibliographystyle{acl_natbib}

\appendix

\section{Runtimes}
\label{sec:runtimes}
Models were run on two Intel Xeon Silver 4210 CPU with 20 cores and 40 threads in total and 187 GiB of system memory.
Runtimes are reported without embedding time, as we wanted a fair comparison between topic models utilizing embeddings from different sentence encoders.

\section{Regression Analysis}
We report coefficients from our regression analysis of the results in Table \ref{tab:regression}.

\begin{table*}[ht]
\centering
\begin{tabular}{lrrrrrr}
\toprule
 & \textbf{Coef.} & \textbf{Std. Err.} & \textbf{t} & \textbf{p-value} & \textbf{[0.025} & \textbf{0.975]} \\
\midrule
Intercept($S^3_{com}$) & 0.6061 & 0.008 & 79.848 & 0.000 & 0.591 & 0.621 \\
$S^3_{ax}$ & -0.0005 & 0.011 & -0.046 & 0.963 & -0.022 & 0.021 \\
$S^3_{ang}$ & -0.0145 & 0.011 & -1.350 & 0.178 & -0.036 & 0.007 \\
FASTopic & -0.0427 & 0.011 & -3.982 & 0.000 & -0.064 & -0.022 \\
ECRTM & -0.0308 & 0.011 & -2.865 & 0.004 & -0.052 & -0.010 \\
Top2Vec & -0.0463 & 0.011 & -4.312 & 0.000 & -0.067 & -0.025 \\
CombinedTM & -0.1204 & 0.011 & -11.215 & 0.000 & -0.141 & -0.099 \\
ZeroShotTM & -0.1170 & 0.011 & -10.901 & 0.000 & -0.138 & -0.096 \\
BERTopic & -0.2141 & 0.011 & -19.949 & 0.000 & -0.235 & -0.193 \\
NMF & -0.2221 & 0.011 & -20.690 & 0.000 & -0.243 & -0.201 \\
LDA & -0.2722 & 0.011 & -25.356 & 0.000 & -0.293 & -0.251 \\
\bottomrule
\end{tabular}
\caption{Regression coefficients for each topic modeling method when predicting aggregate interpretability ($\sqrt{\bar{C} \cdot d}$).}
\label{tab:regression}
\end{table*}

\section{Hyperparameters}
\label{sec:hyperparameters}
With all topic models we chose hyperparameters which were either defaults in their respective software packages or used in widely available online resources.
This was motivated by a number of considerations:
\begin{enumerate}
    \item Hyperparameter optimization is computationally expensive and requires an informed choice about metrics to optimize for. We believe that no single metric is sufficient for describing topic quality. We would also like to avoid explicitly optimizing models for metrics we are evaluating on, as the metrics would cease to measure external validity (which is often a concern with e.g. perplexity as a measure of model fit).
    \item Few, if any, systematic investigations have looked into the effects of hyperparameter choices in most topic models, aside from LDA (\citealp{lda_principled_selection_of_hyperparameters}, \citealp{rethinking_lda}), and we therefore believe it is  difficult for practitioners to make informed choices about hyperparameters in topic models.
    \item While it has mainly been shown in other scientific disciplines, there is a known association between high researcher degrees of freedom/flexibility in analysis and false positive results. (\citealp{researcher_degrees_of_freedom}, \citealp{avoid_p_hacking}, \citealp{false_positive_psychology})
    Arbitrarily tweaking hyperparameters in topic models might make results more biased and prone to the researcher's prior expectations.
\end{enumerate}

\subsection{S\textsuperscript{3}}
For \textbf{S\textsuperscript{3}}, we used default scikit-learn parameters for FastICA: \url{https://scikit-learn.org/stable/modules/generated/sklearn.decomposition.FastICA.html}. The algorithm uses parallel estimation, an SVD solver is used for whitening, and the whitening matrix is rescaled to ensure unit variance.

\subsection{Top2Vec and BERTopic}
For \textbf{Top2Vec} and \textbf{BERTopic} models, all UMAP and HDBSCAN hyperparameters were identical to the defaults of the BERTopic package (\url{https://github.com/MaartenGr/BERTopic}).
Specifically, dimensionality reduction is performed using UMAP (\url{https://umap-learn.readthedocs.io/en/latest/}), with the following parameters: n\_neighbors=15; n\_components=5; min\_dist=0.1; metric="cosine".
Clustering is performed with HDBSCAN (\url{https://hdbscan.readthedocs.io/en/latest/index.html}), with the following parameters: min\_cluster\_size=15, metric="euclidean", and cluster\_selection\_method="eom". 

\subsection{ZeroShotTM and ContextualizedTM}
For \textbf{ZeroShotTM} and \textbf{ContextualizedTM} models, the structure of the variational autoencoder was the following. The encoder network consists of two fully connected layers with 100 nodes each and softplus activation followed by dropout. The outputs of this are passed to mean and variance layers (fully connected) of size equal to the number of topics. Batch normalization is applied to their outputs, and variance vectors are exponentiated to enforce positivity.
The decoder network includes a fully connected layer (preceded by dropout) with size equal to the size of the vocabulary and no bias parameter, followed by batch normalization and softmax activation.

The network was identical for ZeroShotTM and ContextualizedTM, the only difference being the input layer. Inputs to the encoder only include sentence embeddings for ZeroShotTM, while these are concatenated with BoW representations for CombinedTM. 
The following parameters were used for training: batch\_size=42; lr=1e-2; betas=(0.9, 0.999); eps=1e-08, weight\_decay=0; dropout=0.1; nr\_epochs=50.

Much of the implementation, along with the default parameters were taken from the Pyro package's ProdLDA tutorial: 
\url{https://pyro.ai/examples/prodlda.html}

\subsection{ECRTM}
We used default hyperparameters for ECRTM from the TopMost package \footnote{\url{https://github.com/bobxwu/topmost}} (en\_units=200, dropout=0.0,  embed\_size=200, beta\_temp=0.2, weight\_loss\_ECR=100.0, sinkhorn\_alpha=20.0,
        sinkhorn\_max\_iter=1000).

The model was trained with a batch size of 200 for 200 epochs with learning rate 0.002.

\subsection{FASTopic}
The default parameters from the FASTopic package\footnote{\url{https://github.com/BobXWu/FASTopic}} were used for training the models.
These were:
DT\_alpha = 3.0,
        TW\_alpha = 2.0,
        theta\_temp = 1.0,
        n\_epochs = 200,
        learning\_rate = 0.002,

\subsection{LDA and NMF}
For both \textbf{LDA} and \textbf{NMF} we used default parameters from the scikit-learn implementation: \url{https://scikit-learn.org/stable/modules/generated/sklearn.decomposition.LatentDirichletAllocation.html} 
and \url{https://scikit-learn.org/stable/modules/generated/sklearn.decomposition.NMF.html} respectively.

\section{Performance per corpus}
\label{sec:performance_per_corpus}

To examine the performance of different models per corpus, consult  Table 3.

\begin{table*}[p!]
\begin{tabular}{cc}
\begin{tabular}{lccc | c}
\toprule

\textbf{Model} & $d$ & $C_{\text{in}}$ & $C_{\text{ex}}$ &  $\sqrt{\bar{C} \cdot d}$ \\

\midrule
20NG\textsubscript{Pre} & & & &\\
\midrule
$S^3_{\text{axi}}$ &0.90 & 0.33 & 0.19 & 0.47\\
$S^3_{\text{ang}}$ &0.96 & 0.32 & 0.18 & 0.48\\
$S^3_{\text{com}}$ &0.93 & 0.33 & 0.19 & 0.48\\
Top2Vec &0.97 & 0.41 & \textbf{0.23} & \underline{0.54}\\
FASTopic &\underline{0.97} & 0.48 & 0.17 & 0.53\\
ECRTM &0.86 & \textbf{0.57} & \underline{0.21} & \textbf{0.54}\\
BERTopic &\textbf{0.98} & 0.16 & 0.19 & 0.39\\
CTM\textsubscript{combined} &0.94 & 0.48 & 0.16 & 0.51\\
CTM\textsubscript{zero-shot} &0.95 & \underline{0.50} & 0.17 & 0.52\\
LDA &0.74 & 0.35 & 0.17 & 0.42\\
NMF &0.72 & 0.38 & 0.15 & 0.41\\
\midrule
BBC News & & & &\\
\midrule
$S^3_{\text{axi}}$ &0.93 & 0.92 & 0.24 & \textbf{0.66}\\
$S^3_{\text{ang}}$ &\underline{0.98} & 0.91 & 0.20 & 0.64\\
$S^3_{\text{com}}$ &0.95 & 0.92 & 0.22 & \underline{0.66}\\
Top2Vec &0.86 & \underline{0.92} & \textbf{0.27} & 0.65\\
FASTopic &\textbf{1.00} & 0.90 & 0.19 & 0.65\\
ECRTM &0.89 & \textbf{0.93} & 0.19 & 0.61\\
BERTopic &0.43 & 0.59 & \underline{0.26} & 0.41\\
CTM\textsubscript{combined} &0.90 & 0.84 & 0.16 & 0.57\\
CTM\textsubscript{zero-shot} &0.83 & 0.83 & 0.18 & 0.56\\
LDA &0.37 & 0.64 & 0.23 & 0.38\\
NMF &0.47 & 0.62 & 0.24 & 0.42\\
\midrule
ArXivML & & & &\\
\midrule
$S^3_{\text{axi}}$ &0.92 & \underline{0.90} & 0.22 & \textbf{0.64}\\
$S^3_{\text{ang}}$ &0.94 & 0.90 & 0.20 & 0.63\\
$S^3_{\text{com}}$ &0.93 & 0.90 & 0.21 & \underline{0.63}\\
Top2Vec &0.55 & 0.82 & \underline{0.22} & 0.46\\
FASTopic &\textbf{1.00} & 0.87 & 0.15 & 0.60\\
ECRTM &\underline{0.95} & \textbf{0.92} & 0.13 & 0.58\\
BERTopic &0.58 & 0.55 & \textbf{0.24} & 0.45\\
CTM\textsubscript{combined} &0.80 & 0.74 & 0.13 & 0.50\\
CTM\textsubscript{zero-shot} &0.74 & 0.74 & 0.13 & 0.48\\
LDA &0.40 & 0.62 & 0.20 & 0.37\\
NMF &0.55 & 0.60 & 0.17 & 0.41\\

\bottomrule
\end{tabular} &

\begin{tabular}{lccc | c}
\hline
\toprule
\textbf{Model} & $d$ & $C_{\text{in}}$ & $C_{\text{ex}}$ &  $\sqrt{\bar{C} \cdot d}$ \\

\midrule
20NG\textsubscript{Raw} & & & &\\
\midrule
$S^3_{\text{axi}}$ &0.98 & 0.43 & \underline{0.29} & \underline{0.59}\\
$S^3_{\text{ang}}$ &\underline{1.00} & 0.42 & 0.26 & 0.58\\
$S^3_{\text{com}}$ &0.99 & 0.43 & 0.28 & \textbf{0.59}\\
Top2Vec &0.76 & 0.41 & \textbf{0.32} & 0.52\\
FASTopic &\textbf{1.00} & \underline{0.49} & 0.19 & 0.55\\
ECRTM &0.99 & \textbf{0.54} & 0.19 & 0.56\\
BERTopic &0.56 & 0.39 & 0.19 & 0.39\\
CTM\textsubscript{combined} &0.89 & 0.41 & 0.18 & 0.49\\
CTM\textsubscript{zero-shot} &0.91 & 0.44 & 0.18 & 0.51\\
LDA &0.38 & 0.41 & 0.24 & 0.35\\
NMF &0.54 & 0.48 & 0.24 & 0.43\\
\midrule
StackExchange & & & &\\
\midrule
$S^3_{\text{axi}}$ &0.98 & 0.34 & \underline{0.31} & \textbf{0.56}\\
$S^3_{\text{ang}}$ &\underline{1.00} & 0.30 & 0.23 & 0.51\\
$S^3_{\text{com}}$ &0.99 & 0.33 & 0.29 & \underline{0.55}\\
Top2Vec &0.76 & \underline{0.41} & \textbf{0.36} & 0.53\\
FASTopic &0.87 & 0.34 & 0.15 & 0.44\\
ECRTM &\textbf{1.00} & \textbf{0.48} & 0.16 & 0.52\\
BERTopic &0.51 & 0.34 & 0.18 & 0.35\\
CTM\textsubscript{combined} &0.92 & 0.34 & 0.16 & 0.46\\
CTM\textsubscript{zero-shot} &0.91 & 0.35 & 0.17 & 0.47\\
LDA &0.38 & 0.34 & 0.21 & 0.32\\
NMF &0.62 & 0.28 & 0.19 & 0.38\\
\midrule
Wiki Medical & & & &\\
\midrule
$S^3_{\text{axi}}$ &0.99 & 0.38 & 0.35 & 0.60\\
$S^3_{\text{ang}}$ &\underline{1.00} & 0.37 & 0.35 & 0.60\\
$S^3_{\text{com}}$ &0.99 & 0.38 & \underline{0.36} & \underline{0.60}\\
Top2Vec &0.93 & \underline{0.41} & \textbf{0.45} & \textbf{0.63}\\
FASTopic &\textbf{1.00} & 0.41 & 0.28 & 0.58\\
ECRTM &0.99 & \textbf{0.44} & 0.30 & 0.60\\
BERTopic &0.60 & 0.21 & 0.25 & 0.36\\
CTM\textsubscript{combined} &0.72 & 0.25 & 0.19 & 0.40\\
CTM\textsubscript{zero-shot} &0.78 & 0.26 & 0.21 & 0.43\\
LDA &0.28 & 0.20 & 0.26 & 0.25\\
NMF &0.40 & 0.16 & 0.23 & 0.28\\

\bottomrule

\end{tabular}

\end{tabular}
\label{tab:per_corpus}
\caption{Different models' mean diversity and coherence on corpora \\
\textit{Only results for sentence transformers are taken into account, not GloVe.}
}
\end{table*}

\section{NPMI Coherence}

Due to its historical significance in topic modeling literature, we also evaluated topic descriptions using the $C_{\text{NPMI}}$ coherence metric, which assigns higher values to topic descriptions which contain words that often co-occur in the corpus.

Both theoretical considerations and experimental evidence cast doubt on the effectiveness of this metric to reasonably evaluate topic models \citep{incoherence_in_coherence}.
Our results also indicate that the models, which, based on the metrics reported in the main body of the paper and our qualitative evaluations,
extract topics of lowest quality, score very high on this metric, while Top2Vec, S\textsuperscript{3}, and FASTopic typically boast very low scores, well in the negatives (see Table 4).
We deem internal word embedding coherence a better metric for evaluating topic models' internal coherence.

\begin{table*}[h]
\resizebox{\textwidth}{!}{
\begin{tabular}{lllllll}
\toprule
\textbf{Model} & 20NG\textsubscript{Pre} & 20NG\textsubscript{Raw} & ArXivML & BBC News & StackExchange & Wiki Medical \\
\midrule
$S^3_{\text{axi}}$ & -0.05 & -0.21 & -0.33 & -0.29 & -0.20 & -0.17 \\
$S^3_{\text{ang}}$ & -0.05 & -0.22 & -0.32 & -0.32 & -0.22 & -0.19 \\
$S^3_{\text{com}}$ & -0.05 & -0.21 & -0.32 & -0.30 & -0.21 & -0.17 \\
$\text{Top2Vec}$ & 0.05 & -0.18 & -0.29 & -0.26 & -0.19 & -0.08 \\
$\text{FASTopic}$ & 0.03 & -0.07 & -0.18 & -0.10 & -0.08 & -0.09 \\
$\text{ECRTM}$ & \underline{0.11} & -0.07 & -0.16 & -0.04 & -0.07 & -0.11 \\
$\text{BERTopic}$ & 0.02 & \underline{0.05} & \underline{-0.02} & \underline{-0.01} & 0.02 & \underline{-0.01} \\
$\text{CTM}_{\text{combined}}$ & 0.09 & -0.04 & -0.06 & -0.03 & 0.04 & -0.07 \\
$\text{CTM}_{\text{zeroshot}}$ & \underline{0.11} & -0.01 & -0.05 & -0.02 & \underline{0.05} & -0.05 \\
$\text{LDA}$ & 0.10 & 0.04 & -0.06 & -0.03 & 0.02 & -0.03 \\
$\text{NMF}$ & \textbf{0.12} & \textbf{0.09} & \textbf{0.01} & \textbf{0.01} & \textbf{0.06} & \textbf{0.01} \\
\bottomrule
\end{tabular}}
\label{tab:c_npmi}
\caption{ Topic descriptions' average NPMI coherence over corpora. }
\end{table*}

\section{Disaggregated results}
\begin{figure*}[htpb!]
    \includegraphics[width=1\linewidth]{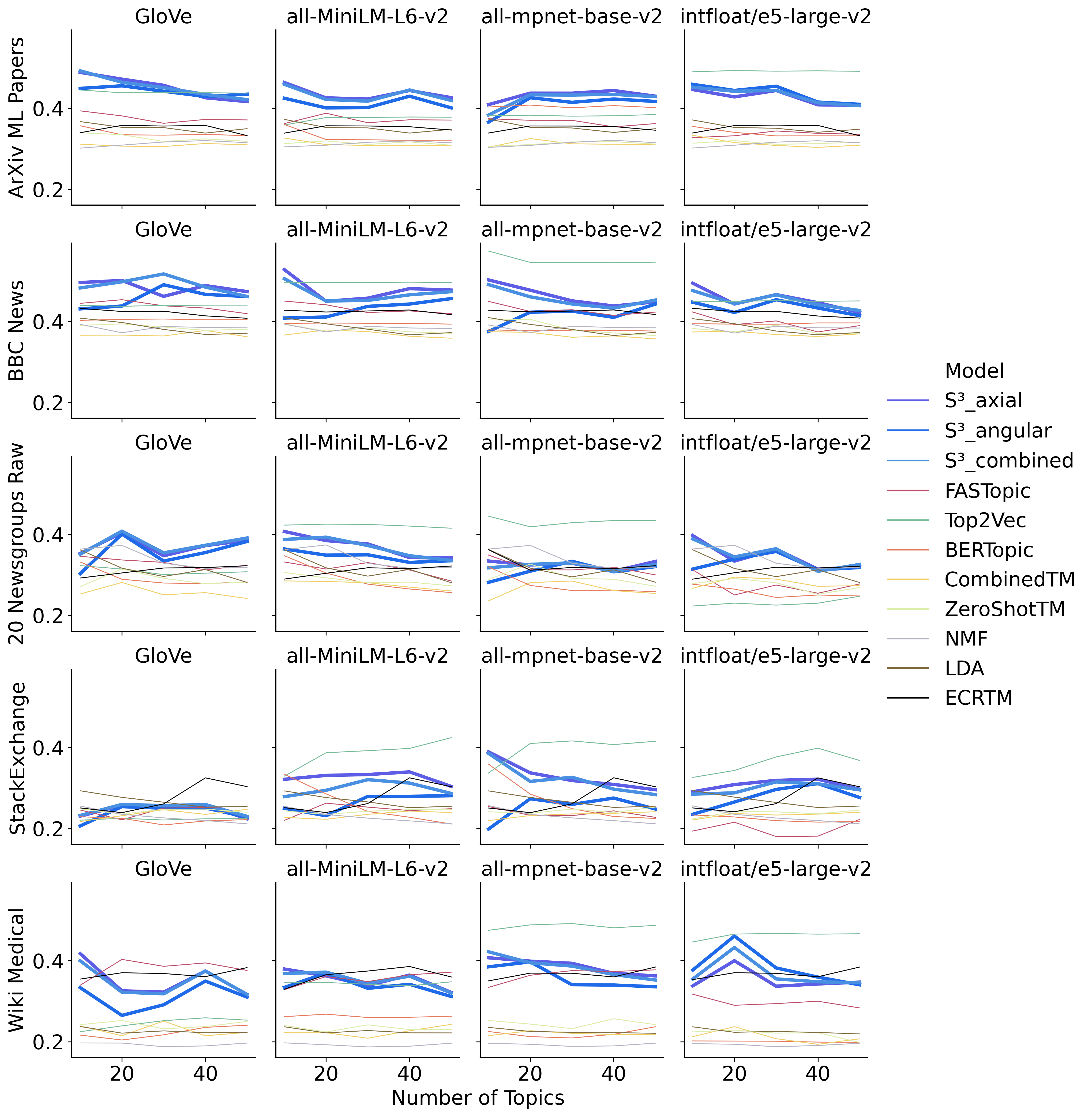}
    \caption{Aggregate coherence scores across all models, datasets, encoders, and numbers of topics, computed as geometric mean of internal and external word embedding coherence}
    \label{fig:disagg_coherence}
\end{figure*}

\begin{figure*}[htpb!]
    \includegraphics[width=1\linewidth]{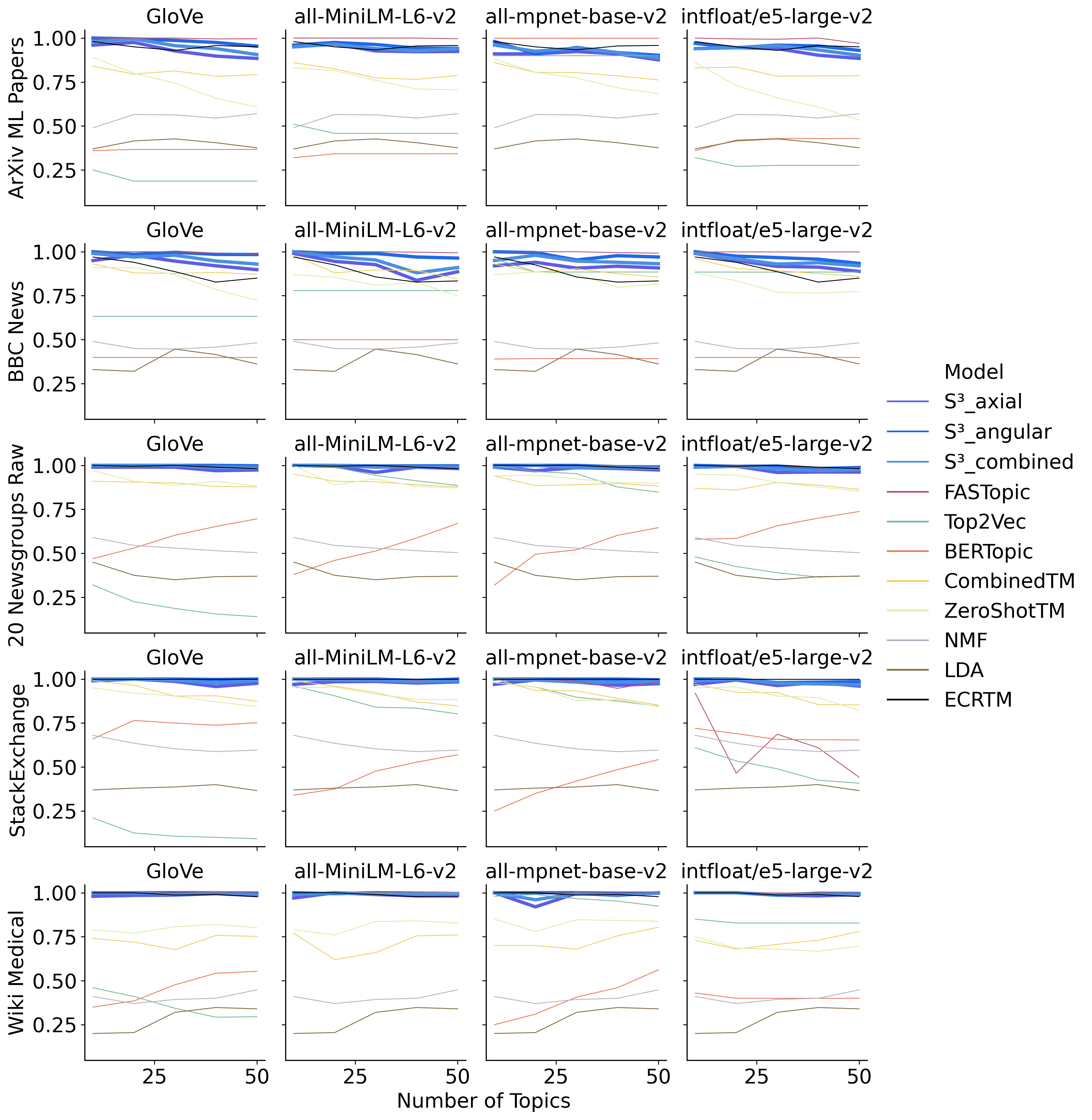}
    \caption{Diversity scores across all models, datasets, encoders, and numbers of topics.}
    \label{fig:disagg_diversity}
\end{figure*}

\begin{figure*}[htpb!]
    \includegraphics[width=1\linewidth]{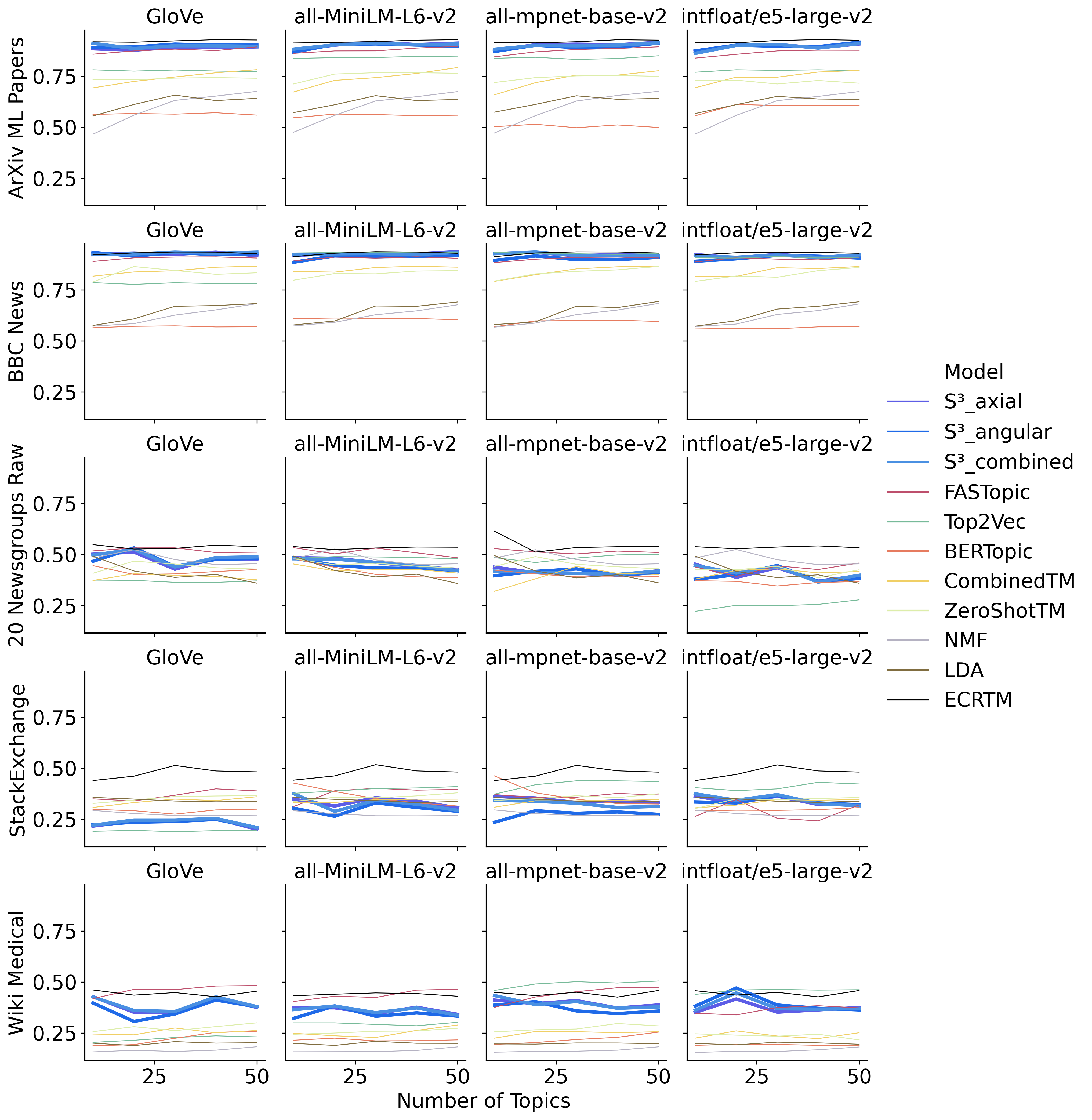}
    \caption{$WEC_{in}$ scores across all models, datasets, encoders, and numbers of topics.}
    \label{fig:disagg_wec_in}
\end{figure*}

\begin{figure*}[htpb!]
    \includegraphics[width=1.0\linewidth]{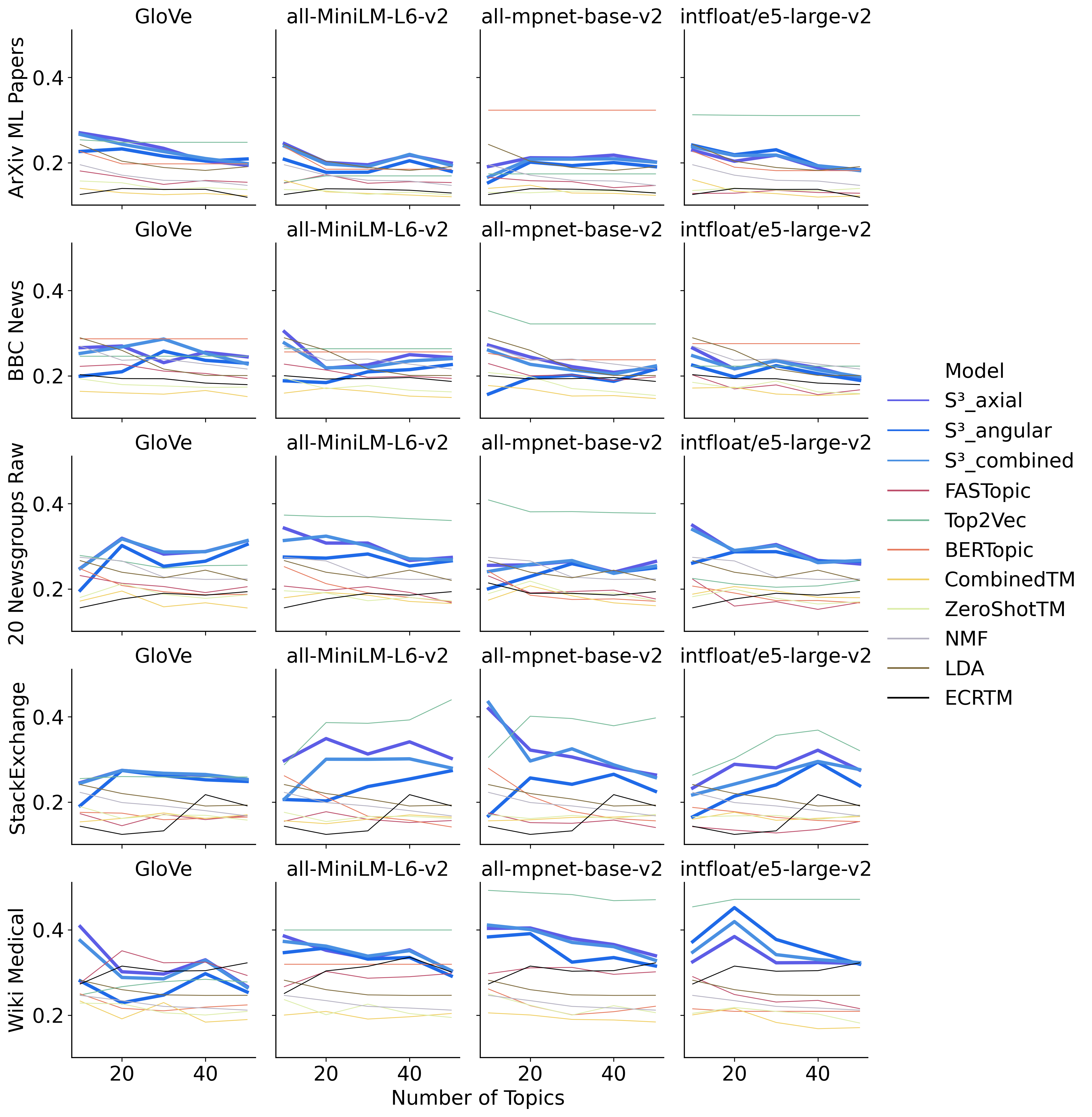}
    \caption{$WEC_{ex}$ scores across all models, datasets, encoders, and numbers of topics.}
    \label{fig:disagg_wec_ex}
\end{figure*}

\begin{figure*}[htpb!]
    \includegraphics[width=1\linewidth]{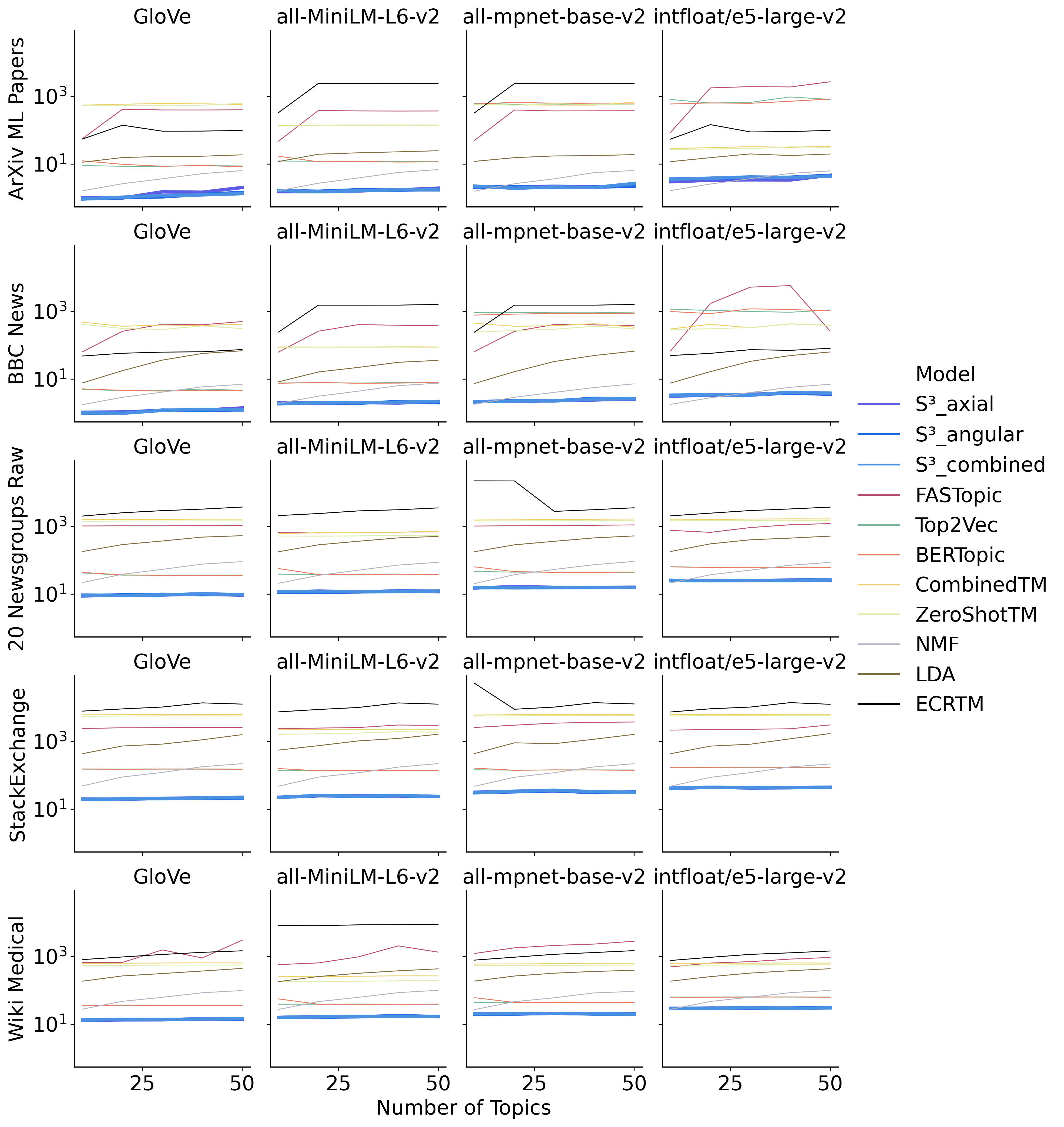}
    \caption{Runtime in seconds for all models, datasets, encoders, and numbers of topics.}
    \label{fig:disagg_runtime}
\end{figure*}

See Figure \ref{fig:disagg_coherence}, \ref{fig:disagg_diversity}, \ref{fig:disagg_wec_in}, \ref{fig:disagg_wec_ex}, \ref{fig:disagg_runtime} for a full overview of coherence, diversity, WEC\textsubscript{ex}, WEC\textsubscript{in} and runtime for all models, datasets, encoders, and numbers of topics.

Note that, on the ArXiv ML Papers dataset and with all-mpnet-base-v2 as en encoder, BERTopic consistently displays a diversity score of 1.0 across all numbers of topics. This is due to BERTopic only estimating a single topic which only contains stop-words. Not only is diversity across topics not a meaningful metric here. A model that only estimates a single topic is simply not usable.

We observe a similar scenario for Top2Vec, which only estimates two topics over the entire ArXiv ML corpus.

\section{Non-alphabetical Terms}
\begin{figure}[h!]
    \includegraphics[width=.92\linewidth]{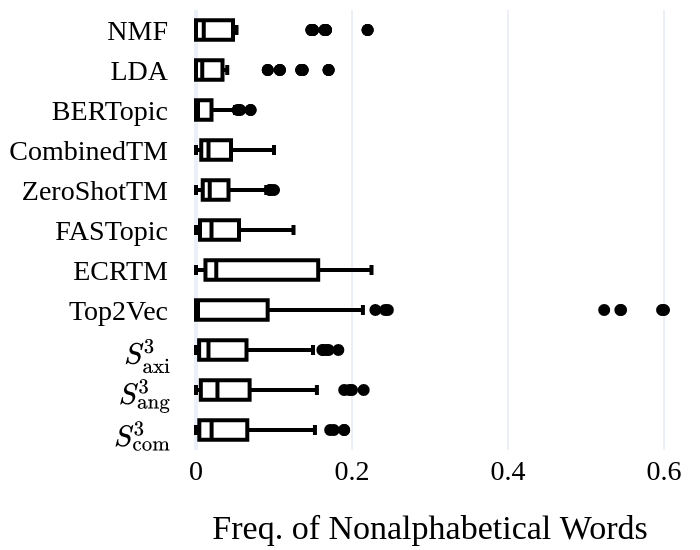}
    \caption{Relative frequency of words containing nonalphabetical characters in topic descriptions}
    \label{fig:nonalphabetical}
\end{figure}

See Figure \ref{fig:nonalphabetical} for information on the proportion of terms containing non-alphabetical characters in topic descriptions.

\section{Licensing}
We release both the CLI and results for quantitative benchmarks (\url{https://github.com/x-tabdeveloping/topic-benchmark}) and the \texttt{Turftopic} Python package (\url{https://github.com/x-tabdeveloping/turftopic}) under the MIT license.

\section{Topic Descriptions from Qualitative Analyses on 20Newsgroups}
\label{sec:topic_desc}
\subsection{No Encoder}\vspace{2mm}
\subsubsection{ECRTM}
\noindent
\textbf{0} - db, ilbm, xloadimage, jpeg, insulation, cec, quantization, xli, wiring, simtel20\\
\textbf{1} - verbeek, billington, cassels, c5ff, nyr, det, bos, guerin, nieuwendyk, ashton\\
\textbf{2} - matusevich, mmatusev, misspelling, refractive, neustaedter, c65oil, donald\_mackie, klosters, benzene, gutmann\\
\textbf{3} - hiv, clinical, screening, nutritional, efficacy, nutrition, metabolic, nonprofit, undergoing, infected\\
\textbf{4} - mafifi, aa824, 1920s, algeria, jihad, 1483500354, jews, refugee, annexed, lehi\\
\textbf{5} - orbit, hayashida, moon, uco, abdkw, reentry, launch, khayash, mars, 4368\\
\textbf{6} - hamer, north1, jlevine, altcit, mavenry, maven, d0i, moyne, cbr900rr, 9733\\
\textbf{7} - jesus, scriptures, god, heresies, redemption, cardenas, affirm, aaronc, holiness, wickedness\\
\textbf{8} - 0d, b8f, 75u, 145, ax, a86, 6um, 1d9, bhj, max\\
\textbf{9} - accounts\_, caratzas, yelena, shahmuradian, microdistrict, sumgait, deposition, aristide, bonner, balcony\\
\textbf{10} - apache, rupin, transferable, airlines, 376, ppd, pong, sonic, yow, listings\\
\textbf{11} - blashephemers, jsn104, sandvik, floated, an030, henceforth, apr22, hela, horne, woodwork\\
\textbf{12} - polonius, insurrection, mi6, neptunium, assualt, semtex, 1715, trailers, firefight, seige\\
\textbf{13} - x11r5, widget, lib, xtwindow, lxaw, xinit, openwindows, xterm, lx11, scrollbar\\
\textbf{14} - relativist, andtbacka, morals, mandtbacka, gullible, \_must\_, kilimanjaro, drewcifer, hens, odwyer\\
\textbf{15} - 5i, ym, 6ei, 0d, \_o, c4, 3j, 7p, 0e, \_5\\
\textbf{16} - encryptions, encryption, rboudrie, escrow, backdoor, 0366, yearwood, fides, trapdoors, unsecure\\
\textbf{17} - coexist, scsi, 120mb, irq5, 80mb, ide, silverlining, powerbooks, card, 33v\\
\textbf{18} - shaz, kokomo, slcs, 46904, rpms, milage, roomy, odometers, hoffmeister, jmh\\
\textbf{19} - doubters, sbp002, announcers, bream, woofing, bitching, hander, umps, mediot, vergolini\\
\subsubsection{LDA}
\noindent
\textbf{0} - that, to, you, of, from, the, and, in, was, on\\
\textbf{1} - to, the, of, com, is, in, that, and, it, be\\
\textbf{2} - the, to, and, is, in, it, of, you, this, for\\
\textbf{3} - is, to, the, in, and, of, that, it, on, for\\
\textbf{4} - the, to, that, is, not, in, it, of, you, and\\
\textbf{5} - b8f, g9v, pl, a86, ax, max, 145, 1d9, 1t, 0t\\
\textbf{6} - to, of, and, is, that, in, the, as, for, was\\
\textbf{7} - 11, 12, 10, 25, 15, 20, 14, 16, 13, 18\\
\textbf{8} - in, that, it, and, of, the, to, they, you, is\\
\textbf{9} - com, hp, netcom, lines, organization, colorado, subject, from, the, db\\
\textbf{10} - nasa, in, of, subject, the, to, from, organization, lines, edu\\
\textbf{11} - of, the, to, writes, edu, com, article, re, is, in\\
\textbf{12} - cx, 0d, \_o, 145, w7, c\_, mv, ah, 34u, lk\\
\textbf{13} - from, that, edu, the, is, of, in, to, it, re\\
\textbf{14} - to, the, of, it, is, and, for, in, with, that\\
\textbf{15} - the, to, of, in, is, and, for, be, are, on\\
\textbf{16} - subject, \_\_\_, apr, 1993, uk, of, lines, \_\_, from, organization\\
\textbf{17} - subject, university, nntp, posting, lines, from, organization, host, edu, com\\
\textbf{18} - for, edu, and, 00, from, subject, of, organization, sale, lines\\
\textbf{19} - is, edu, the, to, from, for, and, dos, it, windows\\
\subsubsection{NMF}
\noindent
\textbf{0} - giz, bhj, pl, 75u, 1t, 3t, max, ax, as, 7ey\\
\textbf{1} - the, to, in, on, was, at, with, this, as, from\\
\textbf{2} - will, this, have, to, be, the, we, and, for, that\\
\textbf{3} - that, the, was, and, they, to, in, it, were, he\\
\textbf{4} - is, of, for, or, and, on, to, the, in, with\\
\textbf{5} - 0t, pl, a86, 34u, 145, b8f, 1d9, wm, bxn, 2tm\\
\textbf{6} - for, in, and, by, of, as, from, are, that, were\\
\textbf{7} - dos, windows, ms, microsoft, or, and, tcp, for, 00, pc\\
\textbf{8} - that, not, the, it, are, of, is, in, this, as\\
\textbf{9} - of, to, you, is, jpeg, and, image, for, it, gif\\
\textbf{10} - 0d, \_o, 145, 34u, 75u, 6um, a86, 6ei, 3t, 0t\\
\textbf{11} - db, and, bh, si, cs, is, al, the, byte, or\\
\textbf{12} - for, edu, com, from, subject, lines, in, organization, article, writes\\
\textbf{13} - 00, 12, 14, 16, 10, 25, 20, 11, 15, 13\\
\textbf{14} - he, of, mr, and, that, we, it, president, you, is\\
\textbf{15} - cx, w7, c\_, t7, uw, ck, lk, hz, w1, mv\\
\textbf{16} - output, file, the, of, to, by, entry, if, for, program\\
\textbf{17} - your, if, it, you, can, that, have, do, are, and\\
\textbf{18} - that, the, jehovah, god, is, he, and, of, lord, his\\
\textbf{19} - a86, b8f, 1d9, g9v, ax, 0d, 75u, b8e, lg, 145\\
\subsection{all-MiniLM-L6-v2}\vspace{2mm}
\subsubsection{$S^3_{ang}$}
\noindent
\textbf{0} - broadcasts, scope, antennas, broadcasters, nfl, wavelength, fox, vhf, radio, announcements\\
\textbf{1} - treatments, medicines, pregnancy, medication, medications, diseases, diagnosed, infections, indications, complications\\
\textbf{2} - theological, christians, devout, theology, christianity, religious, scripture, faith, repentance, pantheism\\
\textbf{3} - discriminatory, homosexual, individuality, oppression, compromises, discrimination, homosexuality, homosexuals, discriminated, openly\\
\textbf{4} - colormap, colormaps, imagemagick, gandalf, imagewriter, photoshop, drawings, images, outlines, bitmap\\
\textbf{5} - foremost, dsu, ssd, ss, data, extensive, extensively, dds, numerous, seanna\\
\textbf{6} - firearms, rifles, gun, armed, guns, ammunition, ammo, firearm, handgun, shooting\\
\textbf{7} - configurable, constants, configurations, parameters, selects, configuration, configuring, calculate, requirements, simultaneous\\
\textbf{8} - unomaha, h2, alabama, mo, missouri, grams, malaysia, pork, electron, shellgate\\
\textbf{9} - 1542b, editions, ratified, gaines, doctrines, umbc, gfk39017, publishes, issuing, slated\\
\textbf{10} - staying, advising, volunteer, delaying, responsiblity, serving, teach, bed, hang, administrators\\
\textbf{11} - macs, macadam, macuser, joystick, mac, apple, maccs, macalester, joysticks, macinnis\\
\textbf{12} - balkans, serbia, turks, balkan, turkish, yugoslavia, armenians, serbian, bosnians, bosnian\\
\textbf{13} - woes, gwu, hsu, bgsu, aft, hs, bore, warmer, standpoint, behold\\
\textbf{14} - pricing, resale, sale, discount, selling, purchase, purchases, buys, dollar, buy\\
\textbf{15} - nasa, planets, spaceflight, spacecraft, interplanetary, space, satellites, solar, moon, earth\\
\textbf{16} - cisc, rdc8, slcs, slac, province, mgr, iscsvax, gfx, mg, bnc\\
\textbf{17} - palestineans, israelites, israelis, israeli, palestinians, israel, zionism, zionist, palestine, aviv\\
\textbf{18} - marked, 167, 44272, 149, 416, passed, 7423, 422, crossed, 7521\\
\textbf{19} - vga, 14400, resolution, monitors, 980, rgb, 1070, resolutions, 640x480, monitor\\
\subsubsection{$S^3_{axi}$}
\noindent
\textbf{0} - antennas, vhf, broadcasts, transmitters, reception, radio, uhf, radios, broadcasters, rf\\
\textbf{1} - diseases, patients, pathology, treatments, medicines, diagnosis, medications, diagnosed, clinics, malpractice\\
\textbf{2} - biblical, scripture, theological, christianity, theology, gospels, scriptures, theologians, christians, devout\\
\textbf{3} - morality, homosexuality, homosexual, heterosexuals, discriminated, discrimination, immoral, discriminatory, homosexuals, heterosexual\\
\textbf{4} - imagewriter, colormaps, xputimage, bitmap, imagemagick, colormap, autocad, bitmaps, animation, canvas\\
\textbf{5} - scholarly, ssd, academic, academia, scientific, academics, sciences, speeds, degrees, benchmarks\\
\textbf{6} - gunshot, guns, pistols, firearms, firearm, handgun, handguns, bullets, gun, gunfire\\
\textbf{7} - configuration, implementations, configurations, configuring, transmitters, angular, parameters, fortran, protocols, ranges\\
\textbf{8} - missouri, mo, khz, aj, albuquerque, malaysia, shellgate, mosque, dodge, jehovah\\
\textbf{9} - umbc, npr, publishes, editions, revelations, manuscripts, gaines, gfk39017, doctrines, 1542b\\
\textbf{10} - volunteer, networking, attend, conferencing, remotely, communicating, occupy, temporarily, broadcast, intervene\\
\textbf{11} - mac, maccs, macs, macuser, logitech, macadam, macworld, macx, macintosh, macinnis\\
\textbf{12} - balkans, bosnians, bosnian, armenian, armenians, turkish, turks, bosnia, armenia, azerbaijani\\
\textbf{13} - campus, attending, ohsu, uc, amherst, bore, osu, aft, coliseum, ucsu\\
\textbf{14} - prices, sale, ebay, deals, pricing, inexpensive, cheap, cost, resale, bargain\\
\textbf{15} - nasa, interplanetary, satellites, spacecraft, astronomy, astronomers, solar, astronomical, astronauts, spaceflight\\
\textbf{16} - pcx, providers, quebec, saskatchewan, provincial, province, lausanne, breton, provider, pcf\\
\textbf{17} - israeli, israelites, aviv, israelis, palestinian, knesset, gaza, palestinians, palestineans, palestine\\
\textbf{18} - 1993apr23, 1993apr24, 1993apr27, 1993apr25, 1993apr21, 1993apr22, 1993apr30, 1993apr17, 1993apr06, 1993apr18\\
\textbf{19} - rgb, monitors, 640x480, monitor, displays, vga, 1070, pixels, resolution, 60hz\\
\subsubsection{$S^3_{com}$}
\noindent
\textbf{0} - reception, radio, antennas, broadcasts, vhf, broadcasters, uhf, transmitters, radios, antenna\\
\textbf{1} - medicines, patients, diseases, treatments, pathology, medications, malpractice, treatment, diagnosis, diagnosed\\
\textbf{2} - biblical, theological, theology, christians, christianity, devout, scripture, scriptures, holiness, gospels\\
\textbf{3} - homosexuals, homosexuality, heterosexuals, discriminated, discrimination, discriminatory, homosexual, immoral, oppression, heterosexual\\
\textbf{4} - bitmaps, xputimage, drawings, photoshop, colormap, colormaps, imagewriter, imagemagick, bitmap, shading\\
\textbf{5} - scholarship, degrees, academia, ssd, scholarly, academic, faculty, academics, phds, doctoral\\
\textbf{6} - pistols, handgun, guns, gun, firearms, firearm, handguns, rifles, bullets, ammo\\
\textbf{7} - configurations, configuration, configuring, ranges, parameters, angular, implementations, requirements, configure, implementation\\
\textbf{8} - missouri, mo, malaysia, khz, jehovah, unomaha, aj, alabama, shellgate, mow\\
\textbf{9} - ratified, revelations, publishes, editions, gaines, doctrines, umbc, gfk39017, 1542b, npr\\
\textbf{10} - intervene, temporarily, volunteer, networking, occupy, attend, conferencing, serving, remotely, communicating\\
\textbf{11} - macx, macuser, macintosh, maccs, macalester, macadam, macs, mac, macworld, logitech\\
\textbf{12} - bosnians, armenians, armenian, bosnian, balkans, turkish, bosnia, turks, serbia, balkan\\
\textbf{13} - ohsu, campus, aft, bore, amherst, osu, marched, attending, bsu, hsu\\
\textbf{14} - cost, deals, pricing, prices, sale, resale, bargain, cheap, ebay, auction\\
\textbf{15} - nasa, satellites, spacecraft, solar, interplanetary, astronomy, astronomers, astronomical, spaceflight, space\\
\textbf{16} - gfx, providers, lausanne, quebec, province, provincial, saskatchewan, pcx, breton, pcf\\
\textbf{17} - aviv, israelites, israelis, israeli, palestinians, palestineans, palestinian, knesset, palestine, gaza\\
\textbf{18} - 1993apr17, 1993apr25, 1993apr27, 1993apr23, 1993apr22, 1993apr21, 1993apr24, 7521, 1993apr30, 422\\
\textbf{19} - vga, monitors, monitor, 1070, resolution, 640x480, rgb, displays, 980, resolutions\\
\subsubsection{BERTopic}
\noindent
\textbf{0} - it, you, is, car, and, in, the, to, com, of\\
\textbf{1} - you, is, to, that, the, it, of, and, in, they\\
\textbf{2} - in, it, to, is, that, and, the, msg, of, this\\
\textbf{3} - the, to, is, it, window, with, from, and, for, in\\
\textbf{4} - the, to, for, and, is, it, edu, of, from, in\\
\textbf{5} - the, to, for, it, of, radar, and, in, is, you\\
\textbf{6} - of, medical, cancer, to, the, is, in, eye, and, circumcision\\
\textbf{7} - 00, sale, wolverine, for, 50, 10, 1st, comics, edu, appears\\
\textbf{8} - space, to, of, the, and, is, in, for, it, that\\
\textbf{9} - science, of, is, the, that, to, in, and, it, not\\
\textbf{10} - g9v, ax, max, 145, pl, a86, b8f, 1d9, 34u, 1t\\
\textbf{11} - the, fan, cpu, cooling, heat, power, on, towers, water, is\\
\textbf{12} - gas, the, oil, it, in, com, to, edu, you, my\\
\textbf{13} - the, insurance, in, care, to, of, private, health, and, that\\
\textbf{14} - wax, chain, duct, it, adhesive, solvent, the, and, plastic, from\\
\textbf{15} - wire, wiring, neutral, ground, the, outlets, outlet, grounding, is, to\\
\textbf{16} - discharge, the, battery, acid, temperature, batteries, lead, concrete, and, is\\
\textbf{17} - that, the, to, in, is, and, of, not, you, god\\
\textbf{18} - int, buttons, style, for, arcade, joystick, joysticks, port, edu, controls\\
\textbf{19} - and, in, game, he, the, to, edu, of, that, was\\
\subsubsection{CombinedTM}
\noindent
\textbf{0} - regards, guidelines, thank, considering, obtain, responses, wanted, legislation, thanks, respond\\
\textbf{1} - buf, posting, pts, van, andrew, columbia, 12, host, 21, pit\\
\textbf{2} - remarks, eye, turning, procedure, examine, abandoned, kinds, requirement, followed, increases\\
\textbf{3} - my, get, car, just, don, me, out, like, ve, re\\
\textbf{4} - maintain, remarks, differently, kinds, examine, remains, precisely, duty, feels, consensus\\
\textbf{5} - encryption, chip, clipper, escrow, com, des, netcom, keys, nsa, key\\
\textbf{6} - cramer, gay, arab, arabs, policy, israeli, optilink, clayton, virginia, jake\\
\textbf{7} - christians, not, god, christ, why, jesus, christian, say, believe, does\\
\textbf{8} - interested, australia, ac, de, morgan, tony, co, demon, uk, advance\\
\textbf{9} - nasa, jpl, orbit, henry, gov, moon, space, shuttle, hst, alaska\\
\textbf{10} - you, is, it, to, ax, not, if, jpeg, that, be\\
\textbf{11} - handled, award, kinds, heavily, appeared, superior, offered, guidelines, conditions, typically\\
\textbf{12} - 145, ax, 0d, \_o, a86, mk, m3, mp, 0g, mm\\
\textbf{13} - dos, drive, card, pc, mac, scsi, ibm, drives, disk, bus\\
\textbf{14} - in, for, on, to, of, and, the, is, be, or\\
\textbf{15} - x11r5, window, motif, manager, application, widget, xterm, usr, visual, lib\\
\textbf{16} - players, game, teams, baseball, ca, team, last, nhl, year, player\\
\textbf{17} - everywhere, demand, prepared, repeated, appeared, shut, exception, destroy, examine, meet\\
\textbf{18} - and, he, that, to, the, they, of, was, in, were\\
\textbf{19} - georgia, uga, gordon, banks, athens, ai, intelligence, geb, keith, artificial\\
\subsubsection{FASTopic}
\noindent
\textbf{0} - x11r5, font, mouse, ini, dialog, icon, cursor, xdm, xpert, ctrl\\
\textbf{1} - guy, gets, hit, insurance, average, msg, clutch, low, 1993apr15, gary\\
\textbf{2} - donation, hirama, angmar, cosmo, kou, hiramb, hiram, unmoderated, dexter, alfalfa\\
\textbf{3} - dod, dog, helmet, riding, bike, radar, ride, cop, infante, egreen\\
\textbf{4} - escrow, keys, secure, nsa, encryption, crypto, amendment, clipper, pgp, enforcement\\
\textbf{5} - armenian, turkish, president, armenians, rights, government, gun, armenia, states, united\\
\textbf{6} - arab, arabs, israeli, israel, koresh, waco, atf, compound, batf, muslims\\
\textbf{7} - temperature, battery, moon, water, heat, kelvin, batteries, cooling, nsmca, mksol\\
\textbf{8} - bible, god, jesus, christ, christians, christian, church, faith, christianity, sin\\
\textbf{9} - cunyvm, gsh7w, hennessy, rosicrucian, sirach, manuscripts, ceremonies, petch, ruler, thyagi\\
\textbf{10} - zoology, sky, photography, higgins, henry, alaska, pluto, krillean, relay, pgf\\
\textbf{11} - pcx, printer, ink, drexel, print, bubblejet, deskjet, laserjet, diablo, ghostscript\\
\textbf{12} - players, team, hockey, season, nhl, teams, league, games, detroit, espn\\
\textbf{13} - scsi, controller, mhz, motherboard, simms, pin, ide, floppy, bios, drives\\
\textbf{14} - file, windows, image, dos, graphics, ftp, space, files, server, color\\
\textbf{15} - swing, uniforms, dodger, biochem, roush, ball, canseco, bchm, umpires, octopus\\
\textbf{16} - pl, max, 145, ax, b8f, g9v, a86, db, 1d9, 0d\\
\textbf{17} - audio, shipping, sale, hst, amp, forsale, condition, offer, asking, wolverine\\
\textbf{18} - cancer, geb, pain, noring, doctor, food, dyer, foods, banks, gordon\\
\textbf{19} - ford, cars, oil, miles, bmw, engine, honda, wheel, dealer, rear\\
\subsubsection{Top2Vec}
\noindent
\textbf{0} - batting, outfielder, baseman, fielder, shortstops, shortstop, inning, pitching, pitchers, hitters\\
\textbf{1} - bosnians, turks, bosnia, genocide, armenians, kurdish, armenian, armenia, balkans, mustafa\\
\textbf{2} - israelis, gaza, zionism, zionists, israeli, palestinians, zionist, knesset, palestineans, palestinian\\
\textbf{3} - bruins, canadiens, nhl, sabres, hockey, canucks, puck, goaltenders, oilers, leafs\\
\textbf{4} - patients, diagnosed, illnesses, diseases, diagnosis, cochrane, fda, disease, treatments, diagnose\\
\textbf{5} - misrepresentation, policy, administrations, proponents, taxpayer, governmental, subsidies, policies, lobbying, misinterpretation\\
\textbf{6} - pistols, unconstitutional, firearm, handgun, firearms, nra, militia, enforcement, handguns, guns\\
\textbf{7} - firing, casualties, gunmen, perpetrators, prosecute, retaliation, assassination, bombed, fires, eyewitness\\
\textbf{8} - nsa, encryption, encryptions, wiretapping, espionage, cryptosystems, spying, cryptosystem, cryptology, eavesdropping\\
\textbf{9} - motorcycling, motorcycles, bmw, automobile, vehicle, vehicles, suzuki, honda, automotive, automobiles\\
\textbf{10} - atheism, homosexuality, morality, fundamentalists, agnosticism, hypocrisy, skeptics, fundamentalism, fundamentalist, hypocritical\\
\textbf{11} - theological, biblical, testament, christianity, scripture, theology, scriptures, exegesis, devout, theologians\\
\textbf{12} - printer, telecom, printers, teltech, ibmpa, mbunix, pdb059, ibm, netcom, telex\\
\textbf{13} - nasa, spacecraft, spaceflight, astronomy, astronomical, astronomers, astrophysical, interplanetary, solar, galactic\\
\textbf{14} - cds, sale, ebay, pricing, purchase, cd300, purchases, prices, pricey, forsale\\
\textbf{15} - monitors, monitor, displays, vga, radiosity, radios, radar, monitored, oscilloscope, detectors\\
\textbf{16} - disk, megadrive, netware, hdd, disks, harddrive, harddisk, workstation, ssd, os\\
\textbf{17} - imagewriter, bitmaps, xputimage, graphics, colormaps, colormap, bitmap, autocad, pixmaps, imagemagick\\
\textbf{18} - xwindows, x11r3, x11r5, xtwindow, xcreatewindow, xfree86, x11, x86, x11r4, xservers\\
\textbf{19} - ss24x, 1070, 680x0, x86, motherboard, motherboards, processor, processors, xfree86, v064mb9k\\
\subsubsection{ZeroShotTM}
\noindent
\textbf{0} - policy, israel, israeli, gay, cramer, optilink, palestinian, gaza, arab, uucp\\
\textbf{1} - ca, year, score, players, team, better, game, fans, last, season\\
\textbf{2} - asking, sale, offer, interested, shipping, condition, sell, price, offers, forsale\\
\textbf{3} - gun, firearms, batf, fbi, firearm, fire, atf, guns, waco, assault\\
\textbf{4} - engine, dod, nasa, car, bike, cars, shuttle, space, orbit, gov\\
\textbf{5} - so, god, do, believe, because, does, not, atheist, say, why\\
\textbf{6} - hi, thanks, specifically, viewer, advance, shareware, conversion, convert, australia, utility\\
\textbf{7} - 0d, max, ax, g9v, \_o, 145, 75u, mk, a86, b8f\\
\textbf{8} - judges, widely, examine, thank, finding, determine, furthermore, repeated, confusion, remains\\
\textbf{9} - 25, 11, 10, 12, 00, 93, 30, 92, 17, 15\\
\textbf{10} - chip, phone, encryption, clipper, des, key, escrow, pgp, rsa, nsa\\
\textbf{11} - is, the, to, and, for, you, or, be, it, in\\
\textbf{12} - sandvik, promise, kent, newton, rutgers, activities, vice, heaven, satan, athos\\
\textbf{13} - dod, banks, gordon, followed, soon, judges, hospital, geb, treatment, motorcycles\\
\textbf{14} - repeated, judges, kinds, learning, furthermore, letters, sit, differently, wise, examine\\
\textbf{15} - whereas, breast, repeated, judges, examine, becoming, pure, importance, authors, empty\\
\textbf{16} - is, of, in, to, the, and, that, not, are, as\\
\textbf{17} - motif, x11r5, application, lib, usr, window, x11, xterm, manager, openwindows\\
\textbf{18} - in, was, the, and, to, that, they, of, it, we\\
\textbf{19} - scsi, card, bus, drive, board, mac, video, ram, apple, cards\\
\subsection{all-mpnet-base-v2}\vspace{2mm}
\subsubsection{$S^3_{ang}$}
\noindent
\textbf{0} - 7951, charset, 766, 768, 789, 972, 168, 791, transcript, 856\\
\textbf{1} - leafs, canucks, bettman, canadiens, puck, hawerchuk, lidstrom, oilers, sabres, bruins\\
\textbf{2} - vga, resolution, graphics, 640x480, framebuffer, rgb, monochrome, 1280x1024, vram, rendering\\
\textbf{3} - homeopathy, illness, symptoms, placebo, healthy, diseases, medications, illnesses, toxins, poisoning\\
\textbf{4} - cryptophones, crypto, secrecy, encryption, unsecure, crypt, cryptosystems, plaintext, encryptions, cryptographically\\
\textbf{5} - winbench, w7, openwindows, openwin, microsoft, windows, vista, delphi, shareware, os\\
\textbf{6} - kbytes, diagnostics, stability, analyse, 0s, stable, reliably, variables, precision, itself\\
\textbf{7} - fond, gene, mills, sole, 1913, martial, lucas, galen, fabrication, younger\\
\textbf{8} - motherboards, chipset, chipsets, powerpc, motherboard, 50mhz, 68070, 68060, 6700, dell\\
\textbf{9} - price, buying, purchaser, buyer, selling, sale, sells, seller, m\_sells, sellers\\
\textbf{10} - harddisk, megadrive, hdd, sda, ssd, diskettes, harddrive, seagate, smartdrive, sdd\\
\textbf{11} - cc, ccu, hpfcso, cdac, ccs, ccastco, cdx, cosmo, deskjet, 5e\\
\textbf{12} - misleading, allegations, propoganda, misinformation, objections, disapproval, sic, denials, disinformation, censors\\
\textbf{13} - gmc, subaru, acura, lexus, chevrolet, saab, sedan, aftermarket, sedans, vw\\
\textbf{14} - xtappaddtimeout, xremote, xy, xw, xcreatewindow, xsession, x1, xopendisplay, cuz, xmodmap\\
\textbf{15} - 423, 436, 426, 431, 3684, 4368, 433, 3401, 361, 424\\
\textbf{16} - lancs, srsd, sen, rri, seca, sep, sparcs, csiro, sps, mpaul\\
\textbf{17} - watergate, wtc, fbihh, waco, indictment, meltdown, fires, firing, promptly, fire\\
\textbf{18} - suis, mainland, prc, nanaimo, cn, qc, c5tenu, fairbanks, qucdn, eu\\
\textbf{19} - unconstitutional, lawful, constitutionally, constitutional, prohibition, liberally, conservatives, conservative, freedoms, legalization\\
\subsubsection{$S^3_{axi}$}
\noindent
\textbf{0} - fax, modem, 866, caller, transcript, 802, voicemail, 806, 928, 886\\
\textbf{1} - canadiens, oilers, puck, leafs, bruins, sabres, canucks, nhl, hawerchuk, habs\\
\textbf{2} - opengl, vga, graphics, 640x480, 1280x1024, pixels, framebuffer, rgb, rendering, graphical\\
\textbf{3} - epilepsy, medical, toxins, medicines, malpractice, resurection, diseases, homeopathy, poisoning, remedies\\
\textbf{4} - cryptographically, cryptophones, cryptographic, cryptology, encryptions, cryptosystems, eavesdropping, encryption, cryptography, crypt\\
\textbf{5} - openwin, programs, shareware, windows, windows3, microsoft, openwindows, exe, winbench, winword\\
\textbf{6} - physics, computations, hp9000, keyboard, ergonomics, computation, calculators, oscilloscope, diagnostics, analyse\\
\textbf{7} - fabrication, surgical, bio, candida, lachman, yeast, wounds, kawasaki, fungus, biochem\\
\textbf{8} - motherboard, motherboards, chipsets, chipset, 68070, cpus, 66mhz, 68060, powerpc, processors\\
\textbf{9} - prices, sale, resale, buyer, price, priced, pricing, selling, sell, sellers\\
\textbf{10} - ssd, disk, smartdrive, harddisk, hdd, megadrive, seagate, harddrive, disks, prodrive\\
\textbf{11} - hpfcso, deskjet, cartridge, cdx, fossil, winco, cdac, cartridges, xerox, scicom\\
\textbf{12} - newsletter, discrimination, censorship, warnings, disinformation, misinformation, complaints, complaint, editorials, propoganda\\
\textbf{13} - chevrolet, sedan, subaru, sedans, acura, vw, car, toyota, porsche, automobiles\\
\textbf{14} - xtwindow, xcreatewindow, xmodmap, xopendisplay, xsession, xtappaddtimeout, x11, xremote, xservers, xserver\\
\textbf{15} - discrepancy, versions, 4368, inconsistencies, spectrometer, 46904, 382761, repository, 3401, 44272\\
\textbf{16} - usaf, dea, cia, cdc, feds, usgs, sep, compounds, esa, mossad\\
\textbf{17} - napalm, gunfire, explosives, fires, waco, wtc, fbi, firefight, fire, watergate\\
\textbf{18} - arctic, mainland, quebec, soyuz, rusnews, fairbanks, tsn, alberta, nanaimo, vancouver\\
\textbf{19} - legalization, prohibition, guns, constitutionally, freedoms, nra, handguns, constitutional, legalizing, firearms\\
\subsubsection{$S^3_{com}$}
\noindent
\textbf{0} - 928, caller, 866, fax, 808, transcript, 806, 871, butcher, 886\\
\textbf{1} - oilers, sabres, leafs, canadiens, hawerchuk, puck, bruins, habs, canucks, nhl\\
\textbf{2} - 640x480, framebuffer, rgb, vga, opengl, graphics, 1280x1024, resolution, rendering, pixels\\
\textbf{3} - poisoning, malpractice, homeopathy, medical, toxins, illness, diseases, medicines, resurection, disease\\
\textbf{4} - encryptions, cryptophones, encryption, cryptography, eavesdropping, cryptosystems, cryptology, plaintext, crypt, cryptographic\\
\textbf{5} - openwin, openwindows, windows, winbench, microsoft, exe, programs, shareware, windows3, vista\\
\textbf{6} - diagnostics, calculators, oscilloscope, analyse, hp9000, ergonomics, keyboard, physics, analyzing, regression\\
\textbf{7} - wounds, lachman, fabrications, surgical, sole, fabrication, yeast, bio, serge, candida\\
\textbf{8} - chipsets, motherboards, powerpc, motherboard, 68070, chipset, cpus, 66mhz, 68060, processors\\
\textbf{9} - buyer, price, sale, priced, prices, pricing, selling, sellers, sell, seller\\
\textbf{10} - hdd, harddrive, harddisk, seagate, disk, smartdrive, ssd, disks, megadrive, diskettes\\
\textbf{11} - deskjet, cdx, hpfcso, fossil, cartridge, cdac, xerox, ccastco, cartridges, winco\\
\textbf{12} - misinformation, complaint, disinformation, complaints, censorship, propoganda, warnings, allegations, discrimination, censors\\
\textbf{13} - vw, subaru, chevrolet, sedan, porsche, sedans, acura, toyota, car, dealership\\
\textbf{14} - xmodmap, xcreatewindow, xtappaddtimeout, xtwindow, xopendisplay, xsession, xremote, xservers, xserver, x11\\
\textbf{15} - discrepancy, versions, 46904, 382761, 4368, 3401, 3684, 436, inconsistencies, 423\\
\textbf{16} - cdc, dea, usaf, cosar, sen, feds, csiro, compounds, cia, sep\\
\textbf{17} - explosives, fires, gunfire, waco, watergate, wtc, fbi, firefight, napalm, fire\\
\textbf{18} - quebec, nanaimo, arctic, fairbanks, mainland, rusnews, cn, c5tenu, alberta, qucdn\\
\textbf{19} - nra, freedoms, prohibition, constitutionally, legalization, legalizing, handguns, constitutional, unconstitutional, abortions\\
\subsubsection{BERTopic}
\noindent
\textbf{0} - the, in, of, and, is, that, to, you, it, not\\
\textbf{1} - and, is, in, of, be, the, it, to, key, that\\
\textbf{2} - of, to, the, and, is, in, it, that, for, msg\\
\textbf{3} - of, it, car, com, the, and, to, in, on, you\\
\textbf{4} - and, to, the, for, is, it, from, of, edu, with\\
\textbf{5} - battery, lead, acid, concrete, batteries, the, discharge, to, and, is\\
\textbf{6} - the, to, it, cpu, cooling, is, of, fan, heat, on\\
\textbf{7} - the, window, to, mouse, is, it, and, keyboard, in, of\\
\textbf{8} - to, the, of, space, and, on, nasa, in, for, is\\
\textbf{9} - that, you, of, protection, protected, and, copy, is, to, the\\
\textbf{10} - and, is, to, in, he, that, the, game, of, edu\\
\textbf{11} - pl, 145, max, a86, 1d9, g9v, 1t, 34u, ax, b8f\\
\textbf{12} - kirlian, krillean, of, is, photography, the, edu, to, it, eye\\
\textbf{13} - outlets, ground, wiring, neutral, outlet, the, wire, is, to, grounding\\
\textbf{14} - habitable, the, of, to, accelerations, is, oxygen, in, acceleration, planets\\
\textbf{15} - p1, polygon, p2, p3, sphere, algorithm, points, polygons, the, den\\
\textbf{16} - sale, of, for, from, edu, 10, and, 50, 00, 1st\\
\textbf{17} - oort, gamma, bursters, detectors, edu, cloud, of, ray, the, are\\
\textbf{18} - computer, to, the, is, and, hacker, software, of, that, edu\\
\textbf{19} - solvent, mask, solder, boards, adhesive, duct, green, tape, is, used\\
\subsubsection{CombinedTM}
\noindent
\textbf{0} - dos, card, bus, windows, pc, scsi, drive, board, disk, controller\\
\textbf{1} - of, to, the, that, be, is, it, ax, and, not\\
\textbf{2} - 10, offer, 50, 15, 00, new, 20, condition, sale, shipping\\
\textbf{3} - the, is, it, to, that, you, in, and, not, have\\
\textbf{4} - importance, judges, examine, becoming, presence, suit, guidelines, categories, demand, primarily\\
\textbf{5} - policy, clipper, israel, israeli, arab, gaza, arabs, palestinian, tim, lebanon\\
\textbf{6} - zoo, hst, gov, henry, dod, spencer, digex, alaska, billion, pat\\
\textbf{7} - sit, popular, referred, exercise, represents, uniform, primary, unlike, challenge, adult\\
\textbf{8} - not, god, does, christians, your, say, do, atheist, what, believe\\
\textbf{9} - are, be, can, or, is, for, edu, and, mail, system\\
\textbf{10} - doctor, 241, prepared, causing, guest, mouth, occurs, 9760, primarily, controlled\\
\textbf{11} - gordon, cmu, banks, ahl, columbia, buffalo, pitt, espn, gld, cc\\
\textbf{12} - respond, broad, campus, thank, categories, guarantee, advance, thanks, responses, southern\\
\textbf{13} - gun, guns, article, fbi, atf, koresh, fire, batf, writes, waco\\
\textbf{14} - good, car, off, year, better, too, really, up, my, last\\
\textbf{15} - by, was, of, were, and, the, in, from, on, as\\
\textbf{16} - create, window, drawing, ac, x11r5, uk, de, application, xterm, motif\\
\textbf{17} - importance, examine, specifically, primarily, exception, solve, demand, implications, suit, account\\
\textbf{18} - georgia, sandvik, newton, kent, jon, uga, athens, keith, livesey, sgi\\
\textbf{19} - ax, mu, 0d, mj, mm, mf, mp, ca, mx, mo\\
\subsubsection{FASTopic}
\noindent
\textbf{0} - miles, dealer, auto, engine, ford, oil, cars, honda, toyota, mustang\\
\textbf{1} - baseball, team, hockey, nhl, season, players, player, fans, teams, game\\
\textbf{2} - bmw, bnr, cage, shaft, ama, dog, bikes, rider, motorcycle, motorcycles\\
\textbf{3} - signal, voltage, deskjet, circuits, circuit, audio, infrared, krillean, manned, ink\\
\textbf{4} - ide, controller, scsi, motherboard, mhz, bios, bus, simms, isa, ram\\
\textbf{5} - shipping, sale, offer, condition, asking, forsale, sell, selling, comics, wolverine\\
\textbf{6} - maddi, heterosexuals, molestation, angmar, cosmo, rosicrucian, hennessy, petch, foard, elf\\
\textbf{7} - israel, muslim, armenia, israeli, arab, turkish, turks, genocide, turkey, muslims\\
\textbf{8} - stadium, jhunix, journalism, uniforms, gtd597a, umpires, acad, infield, gauss, hispanic\\
\textbf{9} - law, government, were, him, his, he, she, said, her, gun\\
\textbf{10} - maria, muscles, fever, chastity, zisfein, cadre, crohn, migraine, headache, hepatitis\\
\textbf{11} - moon, launch, henry, bike, medical, car, dod, orbit, shuttle, mission\\
\textbf{12} - dialing, advance, rs232, programme, faxes, alee, recipes, hugo, menlo, ccu1\\
\textbf{13} - x11r5, font, xterm, lib, fonts, motif, openwindows, widget, window, xlib\\
\textbf{14} - file, files, dos, graphics, software, drive, image, ftp, disk, pc\\
\textbf{15} - escrow, secure, crypto, des, nsa, clipper, pgp, encryption, keys, algorithm\\
\textbf{16} - jesus, bible, church, god, christ, christianity, faith, christians, christian, moral\\
\textbf{17} - murder, keith, bear, taxes, cult, isc, laws, freedom, caltech, nra\\
\textbf{18} - ax, pl, g9v, a86, b8f, max, 1d9, db, 34u, 0d\\
\textbf{19} - seizures, julie, skndiv, dgbt, spdcc, restaurant, stove, standoff, allergic, glutamate\\
\subsubsection{Top2Vec}
\noindent
\textbf{0} - braves, mlb, pitchers, yankees, rbis, shortstops, mets, phillies, hitters, mattingly\\
\textbf{1} - encryptions, encryption, cryptophones, nsa, eavesdropping, cryptography, wiretapping, wiretaps, cryptology, cryptographic\\
\textbf{2} - malpractice, diagnosis, doses, homeopathy, medical, diagnoses, poisoning, toxins, gastroenterology, biomedical\\
\textbf{3} - zionists, palestinians, holocaust, palestine, hamas, palestineans, zionism, palestinian, gazans, zionist\\
\textbf{4} - satellites, astronauts, astronautics, spaceflight, moonbase, orbiter, spacecraft, nasa, interplanetary, astrophysical\\
\textbf{5} - leafs, goaltenders, bruins, hockey, canadiens, oilers, nhl, sabres, canucks, puck\\
\textbf{6} - 911, terrorism, explosives, militia, conspiracy, terrorists, fbi, hostages, shootings, waco\\
\textbf{7} - repression, agnostics, homosexuality, fundamentalists, fundamentalism, morality, ideology, bigotry, religion, creationism\\
\textbf{8} - firearm, firearms, guns, handguns, militia, pistols, nra, militias, legislation, ammunition\\
\textbf{9} - freeways, bikers, motorcyclist, speeding, braking, motorcycling, motorcyclists, motorcycles, driving, motorcycle\\
\textbf{10} - honda, vehicles, vehicle, v8, suv, chassis, mustang, automobile, automobiles, automotive\\
\textbf{11} - creationism, atheist, agnosticism, creationists, agnostics, athiests, atheistic, atheists, atheism, unbelievers\\
\textbf{12} - christians, gospels, theological, theology, scripture, scriptural, christianity, bible, unbelievers, biblical\\
\textbf{13} - openwin, windows3, win3, os, windows, dos6, win31, msdos, vista, microsoft\\
\textbf{14} - disks, seagate, harddisk, ssd, hdd, harddrive, disk, cdrom, megadrive, diskettes\\
\textbf{15} - postage, contacting, bids, catalogs, mails, reprinted, mailing, email, priced, bidding\\
\textbf{16} - transmitters, vhf, electronics, radios, amplifiers, amplifier, loudspeakers, antennas, radar, rca\\
\textbf{17} - xga, 68070, 680x0, vga, monitors, monitor, 68060, 68020, x2773, gtd597a\\
\textbf{18} - 68060, i486, 68020, hardware, powerpc, hp9000, 68070, chipset, microcomputer, packard\\
\textbf{19} - xserver, xdpyinfo, xopendisplay, xtwindow, xwindows, x11, cadlab, xservers, x11r4, graphical\\
\subsubsection{ZeroShotTM}
\noindent
\textbf{0} - and, in, is, to, the, of, as, be, for, are\\
\textbf{1} - it, you, to, 75u, a86, that, ax, g9v, b8f, 145\\
\textbf{2} - year, players, game, better, last, games, team, score, good, season\\
\textbf{3} - mu, mc, mt, mv, w7, a7, t7, a4, mj, cx\\
\textbf{4} - fire, guns, gun, waco, atf, batf, fbi, firearms, koresh, cramer\\
\textbf{5} - solar, nasa, spacecraft, moon, orbit, shuttle, henry, launch, earth, mission\\
\textbf{6} - doctors, popular, direction, primarily, particularly, fat, supporting, becoming, guarantee, importance\\
\textbf{7} - encryption, clipper, key, chip, des, secret, escrow, nsa, algorithm, phone\\
\textbf{8} - furthermore, primarily, examine, fat, circumstances, remarks, precisely, remains, accurate, repeated\\
\textbf{9} - for, can, edu, graphics, windows, dos, software, image, pc, if\\
\textbf{10} - jews, israel, turkish, uucp, israeli, arab, turks, gaza, war, greece\\
\textbf{11} - 00, st, 25, 18, 10, 12, 75, 30, 20, 50\\
\textbf{12} - dod, bmw, car, ride, miles, cars, engine, oil, bike, riding\\
\textbf{13} - ide, bus, drive, video, scsi, board, drives, card, mac, apple\\
\textbf{14} - keith, sgi, activities, jon, newton, sandvik, atheist, vice, mathew, rutgers\\
\textbf{15} - condition, offer, sale, selling, interested, asking, sell, offers, excellent, forsale\\
\textbf{16} - god, why, do, not, believe, your, christians, say, beliefs, what\\
\textbf{17} - to, it, in, that, and, the, was, they, of, we\\
\textbf{18} - event, x11r5, xterm, create, window, xlib, lib, application, usr, draw\\
\textbf{19} - worst, chem, espn, gatech, utoronto, alchemy, gerald, coverage, beat, sas\\
\subsection{average\_word\_embeddings\_glove.6B.300d}\vspace{2mm}
\subsubsection{$S^3_{ang}$}
\noindent
\textbf{0} - service, sms, forwarding, email, hotline, fax, messages, telephone, isp, compuserve\\
\textbf{1} - kgb, amos, storing, bits, slices, bytes, decode, mpeg, manipulate, processing\\
\textbf{2} - obligations, financing, guarantee, guarantees, ensure, ensuring, funds, impose, funding, agreements\\
\textbf{3} - asked, wait, reply, replied, hillary, week, answer, clinton, asking, afterward\\
\textbf{4} - criminal, robbery, charged, assault, suspects, officers, custody, guilty, police, convicted\\
\textbf{5} - windows, microsoft, xp, wordperfect, dos, os, linux, vista, gui, solaris\\
\textbf{6} - wisconsin, campus, college, pennsylvania, university, school, georgetown, connecticut, schools, massachusetts\\
\textbf{7} - hitter, inning, innings, batters, rbi, batted, rbis, shortstop, homers, homer\\
\textbf{8} - afterall, 2131, \_\_\_\_\_\_\_\_\_\_\_\_\_, halat, qh, cdx, yf, yl, bri, p7\\
\textbf{9} - slang, offending, labeled, insulting, cheat, insult, jew, offended, dilemma, choice\\
\textbf{10} - subjective, relevance, implications, contrary, perception, motivation, theories, perspective, attitudes, relate\\
\textbf{11} - jesus, faith, christ, apostles, divine, messiah, resurrection, teachings, god, holy\\
\textbf{12} - fault, responds, behaves, bent, activate, activated, rotor, screw, backwards, cog\\
\textbf{13} - flag, honor, banner, fair, freedom, xyz, independent, patriot, rewrite, stands\\
\textbf{14} - communist, occupation, uprising, liberation, lebanon, turkish, palestinian, conflict, serbs, moslem\\
\textbf{15} - cpu, scsi, pci, motherboard, processor, chipset, pentium, cpus, peripherals, motherboards\\
\textbf{16} - finals, canadiens, devils, goaltender, nhl, canucks, bruins, penguins, leafs, flyers\\
\textbf{17} - \_5, 9\_, 6\_, 0\_, 1\_, 2\_, \_6, 7\_, \_0, \_4\\
\textbf{18} - vice, mcdonald, ceo, founder, motorola, corp, associates, ho, lo, packard\\
\textbf{19} - stuart, julian, clarke, allan, gilbert, gregory, bernard, samuel, jerome, herbert\\
\subsubsection{$S^3_{axi}$}
\noindent
\textbf{0} - telephone, usenet, subscribers, sms, fax, mail, phone, email, messages, hotline\\
\textbf{1} - slices, bytes, cake, mpeg, unsecure, ah, bby, keystrokes, pixels, megabytes\\
\textbf{2} - subsidies, concessions, funds, taxes, tax, treasury, exemption, guarantees, impose, legislation\\
\textbf{3} - reply, minutes, wait, appointment, yeltsin, gmt, minister, briefing, request, replied\\
\textbf{4} - felony, firearms, firearm, handguns, police, handgun, convicted, assault, gun, robbery\\
\textbf{5} - os, solaris, windows, linux, wordperfect, netware, unix, dos, xp, microsoft\\
\textbf{6} - school, college, campus, graduate, polytechnic, university, seminary, undergraduate, faculty, pennsylvania\\
\textbf{7} - batters, hitter, homers, rbi, rbis, inning, innings, batted, baseman, fastball\\
\textbf{8} - gnv, \_\_\_\_\_\_\_\_\_\_, 0w, \_\_\_\_\_\_\_\_\_\_\_\_\_, someones, 0m, yw, \_\_\_\_\_\_\_\_\_, multisync, tix\\
\textbf{9} - cheese, shirts, leather, racist, zf, smoked, shirt, cream, paste, fascist\\
\textbf{10} - empirical, cognitive, mathematical, subjective, scientific, theories, theoretical, factual, psychology, methodology\\
\textbf{11} - god, jesus, christ, baptism, divine, church, communion, worship, apostles, anglican\\
\textbf{12} - throttle, algorithm, fault, vectors, smtp, grounder, rotor, axis, chord, tcp\\
\textbf{13} - xyz, rgb, space, cameras, pixel, format, flag, vga, resolution, creationism\\
\textbf{14} - nagorno, palestinian, ethnic, arab, israeli, ottoman, turkish, kurdish, israel, arabs\\
\textbf{15} - scsi, cpu, pentium, motherboard, megabytes, microprocessor, mhz, vga, ethernet, pci\\
\textbf{16} - canadiens, leafs, hockey, nhl, sabres, blackhawks, canucks, goaltender, goalie, nordiques\\
\textbf{17} - 2\_, \_0, 6\_, 1\_, \_4, 0\_, \_6, \_5, \_f, \_v\\
\textbf{18} - intel, motorola, bt, hi, ceo, 3com, telecommunications, microsystems, founder, chairman\\
\textbf{19} - edward, sir, william, rickc, john, robert, david, joseph, henry, herbert\\
\subsubsection{$S^3_{com}$}
\noindent
\textbf{0} - subscribers, sms, telephone, email, usenet, mail, fax, messages, hotline, phone\\
\textbf{1} - manipulate, keystrokes, cake, slices, bytes, peanuts, unsecure, bby, mpeg, bits\\
\textbf{2} - guarantee, subsidies, concessions, funds, tax, guarantees, obligations, impose, provisions, legislation\\
\textbf{3} - briefing, reply, wait, minutes, replied, reporters, yeltsin, session, gmt, asked\\
\textbf{4} - firearms, firearm, felony, police, assault, handgun, handguns, convicted, robbery, gun\\
\textbf{5} - xp, os, wordperfect, unix, windows, linux, dos, microsoft, solaris, netware\\
\textbf{6} - faculty, college, university, campus, school, polytechnic, pennsylvania, undergraduate, graduate, seminary\\
\textbf{7} - batters, batted, rbis, baseman, hitter, homers, rbi, inning, innings, shortstop\\
\textbf{8} - gnv, 2128, \_\_\_\_\_\_\_\_\_\_\_\_\_, someones, 0m, \_\_\_\_\_\_\_\_\_, \_\_\_\_\_\_\_\_\_\_, 0w, yw, 4o\\
\textbf{9} - slang, smoked, shirts, leather, racist, insulting, cheese, shirt, zf, paste\\
\textbf{10} - theories, methodology, evolution, theoretical, factual, cognitive, statistics, empirical, subjective, iq\\
\textbf{11} - christ, jesus, god, apostles, baptism, worship, church, divine, communion, holy\\
\textbf{12} - tcp, smtp, vectors, rotor, axis, fault, throttle, algorithm, activated, function\\
\textbf{13} - creationism, banner, flag, flags, rgb, xyz, format, freedom, rewrite, resolution\\
\textbf{14} - turkish, moslem, kurdish, ottoman, palestinian, israeli, arabs, arab, kurds, palestinians\\
\textbf{15} - scsi, cpu, motherboard, microprocessor, megabytes, pci, pentium, mhz, ethernet, vga\\
\textbf{16} - nhl, canadiens, canucks, leafs, sabres, goaltender, hockey, blackhawks, goalie, penguins\\
\textbf{17} - 1\_, 2\_, \_0, 0\_, \_4, 9\_, 7\_, 6\_, \_6, \_5\\
\textbf{18} - intel, founder, ceo, hi, motorola, bt, ho, telecommunications, microsystems, 3com\\
\textbf{19} - william, edward, stuart, robert, david, henry, gregory, sir, herbert, john\\
\subsubsection{BERTopic}
\noindent
\textbf{0} - is, the, to, of, that, and, in, not, god, it\\
\textbf{1} - they, the, and, of, in, that, were, was, it, to\\
\textbf{2} - from, to, the, edu, for, sale, subject, of, lines, and\\
\textbf{3} - is, of, in, that, the, to, and, it, for, be\\
\textbf{4} - ide, drive, the, to, and, scsi, bus, card, for, is\\
\textbf{5} - is, polygon, the, to, points, color, of, and, in, this\\
\textbf{6} - for, the, is, and, of, jpeg, to, you, file, in\\
\textbf{7} - that, and, the, to, of, in, by, is, israel, were\\
\textbf{8} - of, the, and, in, ground, is, to, it, wire, you\\
\textbf{9} - of, it, to, is, and, on, in, the, for, that\\
\textbf{10} - laser, the, and, deskjet, monitor, hp, is, with, printer, ink\\
\textbf{11} - window, manager, parcplace, the, to, position, berlin, ethan, boulder, accept\\
\textbf{12} - men, sex, of, cramer, sexual, gay, optilink, kinsey, homosexual, the\\
\textbf{13} - towers, nuclear, dept, fossil, water, cooling, plants, steam, the, tower\\
\textbf{14} - the, university, of, professors, beyer, andi, uva, playboy, schools, virginia\\
\textbf{15} - game, the, team, to, and, edu, in, of, he, that\\
\textbf{16} - motion, the, mouse, byu, problem, jumpy, cursor, driver, it, is\\
\textbf{17} - 0d, 0g, db, \_o, bh, output, 6t, 6um, 145, a86\\
\textbf{18} - in, gamma, of, oort, larson, ray, theory, universe, are, the\\
\textbf{19} - to, the, is, of, key, and, it, be, encryption, that\\
\subsubsection{CombinedTM}
\noindent
\textbf{0} - israeli, policy, geb, gordon, gaza, banks, arab, arabs, uci, georgia\\
\textbf{1} - requirement, treat, maintain, compromise, filled, demand, exercise, effectively, expert, unique\\
\textbf{2} - microsoft, video, pc, dos, diamond, mouse, drivers, windows, driver, card\\
\textbf{3} - cramer, nsa, stratus, escrow, netcom, fbi, clipper, secure, crypto, optilink\\
\textbf{4} - 00, 12, 10, 25, 93, 11, 31, 17, 20, 84\\
\textbf{5} - ide, controller, drive, power, hard, drives, scsi, speed, board, apple\\
\textbf{6} - compromise, solve, finding, widely, presence, absence, numerous, supporting, serves, similarly\\
\textbf{7} - accurate, explanation, surely, repeated, regulations, challenge, exercise, protecting, suffer, treat\\
\textbf{8} - examine, cited, regulations, regard, legislation, closely, requirement, ordinary, explained, furthermore\\
\textbf{9} - for, space, edu, technology, mail, gov, industry, nasa, sci, data\\
\textbf{10} - season, game, year, team, players, hockey, games, nhl, ca, teams\\
\textbf{11} - sell, asking, forsale, sale, interested, condition, offer, ohio, items, offers\\
\textbf{12} - that, the, to, you, is, it, this, not, and, be\\
\textbf{13} - x11r5, ac, window, xterm, uk, de, uni, co, event, draw\\
\textbf{14} - regulations, exercise, increases, maintain, founded, presence, combined, ought, daily, fraud\\
\textbf{15} - christian, god, christians, jesus, atheist, atheism, rutgers, bible, christ, christianity\\
\textbf{16} - think, out, don, just, my, me, would, they, have, re\\
\textbf{17} - is, ax, that, g9v, to, b8f, a86, it, max, of\\
\textbf{18} - ride, dod, spencer, henry, bike, bmw, zoo, alaska, bnr, hst\\
\textbf{19} - to, the, of, in, was, and, that, on, as, they\\
\subsubsection{FASTopic}
\noindent
\textbf{0} - armenia, israeli, arab, muslim, muslims, turks, turkish, israel, turkey, genocide\\
\textbf{1} - obo, condition, forsale, shipping, sale, offer, vhs, bubblejet, cod, sega\\
\textbf{2} - stratus, atf, fbi, batf, sw, waco, cdt, gas, koresh, feds\\
\textbf{3} - splitting, indyvax, concave, clockwise, vulcan, todamhyp, virginity, perpendicular, krillean, sammons\\
\textbf{4} - pit, db, det, 55, 0d, \_o, 00, 1st, la, van\\
\textbf{5} - children, said, our, president, their, we, were, gun, people, she\\
\textbf{6} - christ, christians, christianity, sin, faith, jesus, bible, christian, atheists, god\\
\textbf{7} - b8f, giz, bhj, max, ax, 1d9, pl, g9v, wm, 1t\\
\textbf{8} - motherboard, mhz, simms, isa, modem, gateway, quadra, pin, hd, monitor\\
\textbf{9} - hockey, players, season, league, game, nhl, baseball, games, team, teams\\
\textbf{10} - geb, clayton, banks, cramer, optilink, gordon, gay, kaldis, n3jxp, pitt\\
\textbf{11} - andresen, rogoff, vb30, stamber, broadcasters, dwarner, jrogoff, drm6640, lafibm, logistician\\
\textbf{12} - cview, dxf, tu, dresden, pov, radiosity, shading, lilley, raytracing, louray\\
\textbf{13} - toyota, integra, yamaha, bmw, honda, convertible, liter, opel, audi, nissan\\
\textbf{14} - battery, dod, dealer, bike, ride, buy, cars, engine, car, ohio\\
\textbf{15} - healta, nicho, bcci, rosicrucian, timmbake, hausmann, maddi, tammy, timmons, khan0095\\
\textbf{16} - pc, files, file, ftp, windows, disk, graphics, scsi, dos, server\\
\textbf{17} - msg, doctor, rind, sinus, dyer, allergic, antibiotics, nsmca, pain, fungal\\
\textbf{18} - dockmaster, angmar, mvanheyn, cosmo, laudicina, kasey, nren, hupp, vanheyningen, cleartext\\
\textbf{19} - encryption, escrow, privacy, des, launch, clipper, moon, orbit, shuttle, space\\
\subsubsection{Top2Vec}
\noindent
\textbf{0} - way, come, time, going, got, game, coming, \_know\_, play, know\\
\textbf{1} - way, taken, saying, people, time, come, fact, say, country, government\\
\textbf{2} - fact, actually, way, make, example, need, come, know, kind, \_know\_\\
\textbf{3} - fact, come, want, people, saying, time, say, make, way, believe\\
\textbf{4} - believe, \_know\_, fact, come, actually, know, way, kind, thought, think\\
\textbf{5} - way, come, time, make, actually, example, fact, instead, need, kind\\
\textbf{6} - come, going, actually, way, time, know, kind, like, make, instead\\
\textbf{7} - make, fact, way, need, say, time, want, actually, come, example\\
\textbf{8} - make, need, example, way, instance, use, instead, actually, time, come\\
\textbf{9} - example, actually, way, fact, need, come, use, instance, make, time\\
\textbf{10} - fact, say, \_know\_, way, come, know, actually, going, time, want\\
\textbf{11} - time, fact, addition, instance, come, example, need, way, make, instead\\
\textbf{12} - fact, actually, \_know\_, way, come, know, time, think, say, make\\
\textbf{13} - come, say, \_know\_, need, way, actually, make, know, fact, want\\
\textbf{14} - available, time, addition, 20, example, 10, instead, make, 30, 15\\
\textbf{15} - instead, way, actually, need, example, use, make, using, instance, time\\
\textbf{16} - time, example, computer, need, using, use, available, instance, instead, computers\\
\textbf{17} - computers, need, example, instance, use, computer, hard, using, way, drive\\
\textbf{18} - instance, example, instead, way, need, time, using, make, use, actually\\
\textbf{19} - example, instance, use, available, using, computer, uses, web, information, software\\
\subsubsection{ZeroShotTM}
\noindent
\textbf{0} - year, he, team, good, game, better, players, his, up, player\\
\textbf{1} - motif, library, application, request, widgets, programming, address, 3d, xlib, lcs\\
\textbf{2} - converted, primarily, prepared, previously, maintain, forth, primary, growing, furthermore, eventually\\
\textbf{3} - cars, dod, engine, bike, ride, miles, car, bmw, road, front\\
\textbf{4} - for, is, ax, to, in, the, and, of, as, edu\\
\textbf{5} - espn, columbia, coverage, cmu, cup, sas, buffalo, tonight, stanley, gld\\
\textbf{6} - can, your, be, if, or, you, this, it, use, are\\
\textbf{7} - christians, jesus, believe, god, christian, scripture, christ, christianity, atheist, bible\\
\textbf{8} - card, dos, pc, mouse, windows, bus, memory, drive, video, os\\
\textbf{9} - 12, 30, 10, 11, 92, 00, 15, 25, 20, 17\\
\textbf{10} - anon, maintain, successful, challenge, primarily, eventually, becomes, portion, warn, empty\\
\textbf{11} - optilink, israeli, cramer, gay, israel, arab, arabs, israelis, department, palestinians\\
\textbf{12} - furthermore, legislation, examine, primarily, primary, anon, significantly, authors, differences, refuse\\
\textbf{13} - was, the, to, you, that, it, they, have, we, but\\
\textbf{14} - of, in, as, and, the, their, by, were, from, was\\
\textbf{15} - privacy, clipper, phone, crypto, eff, netcom, secret, escrow, wiretap, nsa\\
\textbf{16} - mo, \_o, c\_, ax, 0d, mj, mf, cx, mp, 0g\\
\textbf{17} - asking, sale, offers, offer, condition, interested, sell, forsale, shipping, items\\
\textbf{18} - spacecraft, launch, nasa, moon, mission, orbit, shuttle, space, gov, earth\\
\textbf{19} - keith, gordon, pitt, vice, surrender, banks, geb, soon, tek, shameful\\
\subsection{intfloat/e5-large-v2}\vspace{2mm}
\subsubsection{$S^3_{ang}$}
\noindent
\textbf{0} - vga, motherboards, powerbooks, hdtv, processors, vlsi, 8bit, chipsets, motherboard, microprocessor\\
\textbf{1} - incendiary, firefight, wackos, waco, baylor, atf, gunfire, texans, koresh, jonestown\\
\textbf{2} - intifada, zionists, bds, zionist, israelis, mossad, palestinians, likud, israeli, hamas\\
\textbf{3} - cryptophones, cryptographically, encryption, encrypting, encryptions, cryptographic, encrypt, encrypted, cryptography, decryption\\
\textbf{4} - inefficient, underestimate, poor, maximize, optimally, thinking, discourage, unreasonable, rationalize, minimize\\
\textbf{5} - o7, rightfully, rightful, fleet, dammit, vehicles, based, deserved, gearbox, hammers\\
\textbf{6} - iup, uiuc, univ, universitaet, universite, iupui, university, universities, professors, uchicago\\
\textbf{7} - nysernet, alternatives, toolkits, newsgroups, interests, organizations, listings, libraries, communities, recommendations\\
\textbf{8} - nist, wiley, kbytes, ito, usgs, informatik, sdio, applelink, integer, isbn\\
\textbf{9} - ballplayers, rbis, hitters, outfielder, mlb, baseman, baseball, inning, basemen, shortstops\\
\textbf{10} - sceptics, insulting, macworld, delusional, laserwriter, overpriced, laughable, doubters, hoax, donald\_mackie\\
\textbf{11} - mcsnet, usenet, fermilab, hepnet, networks, multiuser, smtp, network, servers, protocols\\
\textbf{12} - usrobotics, nriz, networks, micron, graphs, computational, bottleneck, labs, researchers, attackers\\
\textbf{13} - uhf, voltmeter, transmitters, 50mhz, 40mhz, magnetometer, vhf, wiring, 33mhz, transmitter\\
\textbf{14} - astronautics, skylab, satellites, spacecraft, spaceflight, spacelab, astronauts, interplanetary, aerospace, launches\\
\textbf{15} - obligatory, compares, utxvms, comparable, compare, examples, dozens, alternatively, comparatively, compared\\
\textbf{16} - deals, sold, sell, forsale, deal, seller, bargain, sale, offers, buy\\
\textbf{17} - pretty, publicized, colors, prettier, perceptions, beauty, colours, colour, color, attractive\\
\textbf{18} - sponsoring, objections, opposes, editorials, sponsors, licensing, concessions, noncommercial, advertisers, hopefully\\
\textbf{19} - colormap, xputimage, pixmaps, pixmap, colormaps, 320x200x256, 320x200, imagewriter, renderer, xcreatewindow\\
\subsubsection{$S^3_{axi}$}
\noindent
\textbf{0} - processors, motherboard, motherboards, hdtv, vram, 8bit, microprocessor, vga, semiconductors, powerbooks\\
\textbf{1} - texans, firefight, gunfire, gunshot, baylor, koresh, atf, massacred, waco, exploded\\
\textbf{2} - zionists, israelis, israeli, intifada, zionist, israel, palestinians, likud, palestinian, isreal\\
\textbf{3} - encryptions, encrypting, decryption, cryptographically, encrypt, encrypted, cryptophones, encryption, cryptographic, cryptography\\
\textbf{4} - equilibrium, poverty, morality, thinking, speeding, optimally, maximize, crime, optimizing, libertarianism\\
\textbf{5} - vehicle, fleet, gearbox, clarkson, audi, rightfully, cars, rightful, faithful, vehicles\\
\textbf{6} - iup, universite, univ, university, uiuc, universitaet, universities, iupui, uwaterloo, undergraduate\\
\textbf{7} - alternatives, newsgroups, communities, toolkits, militias, recommendations, shotguns, sects, ideologies, prospects\\
\textbf{8} - nist, integer, ucsd, usgs, bike, sfwa, fwd, isbn, ito, distance\\
\textbf{9} - outfielder, mlb, baseball, baseman, ballplayers, inning, batting, rbis, hitters, ballpark\\
\textbf{10} - hoax, insulting, laserwriter, overpriced, delusional, sceptics, arrogance, macworld, macintosh, rejection\\
\textbf{11} - computing, mcsnet, networks, fermilab, usenet, hepnet, smtp, atheists, multiuser, telepathy\\
\textbf{12} - microelectronics, bottleneck, supercomputing, nriz, graphs, computational, networks, attackers, researchers, graph\\
\textbf{13} - transmitters, voltmeter, magnetometer, 40mhz, transmitter, voltages, 50mhz, 25mhz, vhf, 5mhz\\
\textbf{14} - skylab, spaceflight, spacecraft, astronautics, astronauts, nasa, spacelab, starbase, aerospace, interplanetary\\
\textbf{15} - obligatory, utxvms, compares, alternatively, excerpts, comparatively, comparable, dozens, 1993mar31, examples\\
\textbf{16} - bargain, sold, forsale, deal, sell, deals, offers, purchase, sale, selling\\
\textbf{17} - beauty, color, colours, colour, pretty, colors, visuals, photoshop, beautiful, photographic\\
\textbf{18} - proposals, sponsors, policies, lobbying, sponsoring, concessions, editorials, licensing, lawmakers, opposes\\
\textbf{19} - pixmap, colormaps, colormap, rendered, xputimage, renderer, rendering, redraw, graphics, render\\
\subsubsection{$S^3_{com}$}
\noindent
\textbf{0} - microprocessor, powerbooks, processors, motherboard, motherboards, vga, hdtv, 8bit, vram, semiconductors\\
\textbf{1} - waco, atf, koresh, gunfire, firefight, baylor, texans, wackos, massacred, exploded\\
\textbf{2} - zionist, zionists, likud, israel, israelis, intifada, israeli, palestinians, bds, palestinian\\
\textbf{3} - encrypt, encryptions, cryptographically, encryption, encrypting, cryptophones, encrypted, decryption, cryptographic, cryptography\\
\textbf{4} - optimally, maximize, morality, poverty, thinking, stupidity, rationality, optimizing, equilibrium, speeding\\
\textbf{5} - rightful, vehicles, fleet, vehicle, rightfully, cars, gearbox, faithful, o7, based\\
\textbf{6} - uiuc, iup, universitaet, univ, university, uwaterloo, universities, undergraduate, iupui, universite\\
\textbf{7} - recommendations, sects, toolkits, workgroups, militias, alternatives, newsgroups, communities, organizations, ideologies\\
\textbf{8} - integer, sfwa, bike, fwd, dir, isbn, usgs, ito, nist, ucsd\\
\textbf{9} - inning, baseball, outfielder, ballplayers, rbis, mlb, baseman, hitters, batting, pitchers\\
\textbf{10} - delusional, hoax, insulting, sceptics, laserwriter, overpriced, macworld, arrogance, rejection, ignorant\\
\textbf{11} - network, mcsnet, multiuser, usenet, fermilab, computing, smtp, hepnet, networks, protocols\\
\textbf{12} - nriz, graphs, computational, microelectronics, networks, bottleneck, attackers, researchers, micron, labs\\
\textbf{13} - vhf, magnetometer, 25mhz, 40mhz, transmitters, 50mhz, transmitter, voltmeter, voltages, 5mhz\\
\textbf{14} - spacecraft, astronauts, nasa, skylab, spaceflight, astronautics, aerospace, satellites, spacelab, interplanetary\\
\textbf{15} - xloadimage, utxvms, dozens, obligatory, examples, compares, 1993mar31, comparatively, comparable, alternatively\\
\textbf{16} - bargain, forsale, sold, sale, deals, deal, offers, sell, purchase, seller\\
\textbf{17} - colour, colours, beauty, pretty, color, prettier, colors, visuals, perceptions, beautiful\\
\textbf{18} - licensing, concessions, lobbying, opposes, sponsoring, editorials, sponsors, lawmakers, policies, proposals\\
\textbf{19} - colormap, xputimage, rendered, rendering, colormaps, renderer, pixmaps, pixmap, redraw, graphical\\
\subsubsection{BERTopic}
\noindent
\textbf{0} - the, of, to, in, space, it, edu, is, that, and\\
\textbf{1} - is, it, to, the, for, in, and, of, you, on\\
\textbf{2} - that, it, to, in, of, and, the, is, you, they\\
\textbf{3} - the, in, he, game, to, of, and, edu, that, is\\
\textbf{4} - is, to, in, and, of, that, god, the, you, not\\
\textbf{5} - lafayette, lafibm, lowenstein, players, roger, baseball, princeton, edu, jewish, vb30\\
\textbf{6} - it, is, copy, you, the, of, and, xv, to, edu\\
\textbf{7} - to, the, of, and, dos, is, windows, window, it, for\\
\textbf{8} - for, and, the, edu, 00, sale, from, of, to, in\\
\textbf{9} - lead, battery, the, acid, discharge, it, batteries, temperature, concrete, is\\
\textbf{10} - the, neutral, wire, wiring, ground, outlets, outlet, is, grounding, to\\
\textbf{11} - captain, traded, captains, was, he, team, season, the, resigned, striped\\
\textbf{12} - president, that, we, to, of, mr, the, and, myers, he\\
\textbf{13} - phones, dialing, tone, phone, sweden, se, germany, berlin, universal, wall\\
\textbf{14} - and, jpl, the, of, spacecraft, mars, planet, baalke, earth, solar\\
\textbf{15} - gvg47, tek, valley, daily, verse, petch, gvg, chuck, grass, lines\\
\textbf{16} - mithras, the, order, in, rosicrucian, ch981, of, tony, was, and\\
\textbf{17} - graphics, newsgroup, split, group, aspects, comp, engin, this, of, groups\\
\textbf{18} - cpu, hours, 24, off, monitors, on, power, day, monitor, edu\\
\textbf{19} - god, love, the, apple, kent, newton, sandvik, and, malcolm, royalroads\\
\subsubsection{CombinedTM}
\noindent
\textbf{0} - heavily, negative, repeated, suddenly, originally, furthermore, detail, kinds, addressed, supporting\\
\textbf{1} - arabs, killed, arab, israeli, israel, fbi, atf, fire, waco, adam\\
\textbf{2} - drive, scsi, card, drives, dos, pc, bus, controller, ide, ram\\
\textbf{3} - beginning, primarily, sit, repeated, equivalent, offered, importance, increases, kinds, filled\\
\textbf{4} - drawing, lib, de, ac, uk, event, xterm, draw, window, x11r5\\
\textbf{5} - ax, that, to, the, it, is, of, and, you, in\\
\textbf{6} - sandvik, georgia, gordon, banks, newton, kent, geb, keith, rutgers, jon\\
\textbf{7} - and, or, are, edu, be, for, is, can, with, this\\
\textbf{8} - henry, nasa, zoo, hst, shuttle, gov, gamma, jpl, orbit, moon\\
\textbf{9} - 30, 15, 10, 11, 00, 12, 20, sale, 18, 25\\
\textbf{10} - portion, detail, huge, sit, repeated, suggested, allowing, treated, challenge, answered\\
\textbf{11} - that, it, have, the, not, is, to, this, you, they\\
\textbf{12} - escrow, netcom, chip, encryption, key, clipper, des, algorithm, keys, nsa\\
\textbf{13} - off, better, year, down, didn, players, game, out, up, too\\
\textbf{14} - and, in, as, of, by, to, was, from, the, were\\
\textbf{15} - mo, db, w7, md, mt, mj, bh, mx, mv, columbia\\
\textbf{16} - bike, com, oil, bmw, stratus, nec, dod, ride, cb, cars\\
\textbf{17} - portion, forth, examine, detail, repeated, holds, sit, literally, touch, constantly\\
\textbf{18} - believe, do, not, say, christians, jesus, god, atheist, bible, beliefs\\
\textbf{19} - examined, unlike, repeated, drawn, filled, direction, forth, bringing, examine, explanation\\
\subsubsection{FASTopic}
\noindent
\textbf{0} - max, internet, file, email, windows, ax, dos, files, info, graphics\\
\textbf{1} - kingston, balls, brent, brett, williams, plate, mets, jhu, barry, hcf\\
\textbf{2} - bike, dod, engine, advice, mot, miles, bmw, ride, ford, dealer\\
\textbf{3} - israel, gun, jesus, god, christian, evidence, jews, bible, war, religion\\
\textbf{4} - ohio, texas, jason, austin, sold, magnus, selling, portal, steven, demon\\
\textbf{5} - gordon, banks, geb, fsu, covington, theodore, chronic, uga, halat, mcovingt\\
\textbf{6} - melbourne, tmc, wang, xerox, paradox, und, symbol, tut, byu, std\\
\textbf{7} - natural, tek, fred, sw, caltech, theory, thread, thoughts, waste, tank\\
\textbf{8} - uwaterloo, fraser, surgery, acsu, kidney, carnegie, sfu, tickets, mellon, upenn\\
\textbf{9} - bethesda, umass, wpi, hiram, usc, carolina, vhs, bach, handbook, 7000\\
\textbf{10} - nntp, games, team, chip, mike, ca, netcom, game, car, steve\\
\textbf{11} - modem, cpu, drives, ram, serial, monitor, ide, port, scsi, controller\\
\textbf{12} - reserve, moon, mission, station, orbit, henry, shuttle, princeton, uci, digex\\
\textbf{13} - lin, sdsu, ecs, babb, wharton, gandler, wireless, unixg, mcd, fiberglass\\
\textbf{14} - elementary, undercover, 27th, sued, calstate, hela, eis, protein, pbs, 1980s\\
\textbf{15} - blue, sale, 03, 500, 55, runs, la, 200, nj, van\\
\textbf{16} - and, the, in, to, is, of, that, for, it, you\\
\textbf{17} - mrc, 9591, cain, 2178, violet, parachute, claudio, rintintin, mcwilliams, circumference\\
\textbf{18} - clinic, therapies, seizures, spdcc, treating, bih, infj, antibiotics, distress, albicans\\
\textbf{19} - hello, appreciated, greatly, advance, expose, thanx, cam, xpert, sphere, pointers\\
\subsubsection{Top2Vec}
\noindent
\textbf{0} - regards, fyi, prob, huh, info, umm, commented, regarding, hmm, \_o\\
\textbf{1} - dr, medicine, antibiotics, health, info, fyi, diseases, medications, prob, indications\\
\textbf{2} - reply, commented, hmmmm, regarding, umm, info, hmm, fyi, excerpt, also\\
\textbf{3} - info, cryptography, prob, fyi, excerpt, hmm, encryptions, encryption, cryptanalysis, wiretapping\\
\textbf{4} - info, fyi, baseball, commented, prob, regarding, huh, please, anyway, mlb\\
\textbf{5} - fyi, reply, regarding, info, commented, umm, hmmmm, hmm, hmmmmm, anyway\\
\textbf{6} - please, goaltenders, nhl, goaltending, commented, hockey, prob, regarding, fyi, info\\
\textbf{7} - umm, commented, reply, regarding, info, fyi, hmmmmm, hmm, hmmmm, responded\\
\textbf{8} - satellites, astronomy, spacecraft, science, nasa, fyi, aerospace, commented, hmm, prob\\
\textbf{9} - commented, answers, amen, excerpt, christianity, theology, ahem, hmmmm, hmm, reply\\
\textbf{10} - ahem, hmmmm, commented, umm, hmm, info, example, reply, huh, hmmmmm\\
\textbf{11} - deals, please, interested, info, pls, computers, offered, regards, listing, fyi\\
\textbf{12} - electronics, microcontrollers, transistors, microelectronics, microcontroller, fyi, wiring, capacitors, prob, info\\
\textbf{13} - regarding, \_o, interested, from, reply, fyi, please, email, info, pls\\
\textbf{14} - motherboards, microcontrollers, computers, computer, fyi, computing, 44mb, isbn, info, 2\_\\
\textbf{15} - v\_, info, interested, please, regards, motherboards, computer, computers, computing, fyi\\
\textbf{16} - \_o, fyi, computers, windows, 2\_, computing, prob, info, o\_, \_u\\
\textbf{17} - microcontrollers, computing, computers, motherboards, processors, computer, fyi, microprocessor, prob, microelectronics\\
\textbf{18} - examples, graphics, coordinates, o\_, \_o, images, \_u, p\_, example, graphs\\
\textbf{19} - 0\_, \_o, example, fyi, xtappaddtimeout, 2\_, s\_, o\_, \_u, 1\_\\
\subsubsection{ZeroShotTM}
\noindent
\textbf{0} - atheist, god, say, believe, do, not, what, think, why, who\\
\textbf{1} - sternlight, eff, stratus, netcom, sw, david, secret, va, online, digex\\
\textbf{2} - geb, kent, sandvik, newton, keith, banks, gordon, activities, rutgers, catholic\\
\textbf{3} - essentially, thank, differences, delivery, primarily, growing, guarantee, contrast, supporting, hidden\\
\textbf{4} - that, and, in, as, the, to, of, is, for, it\\
\textbf{5} - 6ei, ax, 6um, b8f, 0d, a86, \_o, g9v, 145, 0t\\
\textbf{6} - 17, 30, 00, 20, 12, 25, 10, sale, 15, 27\\
\textbf{7} - were, in, was, they, that, people, we, and, the, there\\
\textbf{8} - team, nhl, ca, season, hockey, cup, espn, game, teams, games\\
\textbf{9} - numerous, requirement, legislation, hb, primarily, deep, pk, aim, repeated, examine\\
\textbf{10} - key, escrow, clipper, gun, government, encryption, enforcement, fbi, des, guns\\
\textbf{11} - or, for, and, is, you, can, edu, are, to, be\\
\textbf{12} - decisions, prepared, leave, schools, sit, curious, spot, twenty, v6, breast\\
\textbf{13} - db, window, x11r5, lib, xterm, motif, xlib, application, tu, usr\\
\textbf{14} - australia, iii, hi, advance, parallel, thanks, directly, monitor, looking, interested\\
\textbf{15} - israel, israeli, gaza, arab, policy, palestinian, un, arabs, peace, palestine\\
\textbf{16} - get, good, my, car, doctor, diet, patients, ve, too, like\\
\textbf{17} - challenge, prepared, daily, popular, harm, abandoned, sleep, repeated, kinds, importance\\
\textbf{18} - card, ide, bus, scsi, drive, board, disk, drives, mb, pc\\
\textbf{19} - solar, moon, spacecraft, dod, launch, henry, orbit, nasa, earth, mission\\

\end{document}